\documentclass[a4paper,fleqn]{cas-dc}

\usepackage[numbers]{natbib}
\usepackage{amsmath,amssymb,amsfonts}
\usepackage{algorithmic}
\usepackage{graphicx}
\usepackage{multirow}
\usepackage{algorithm,algorithmic}
\usepackage{hyperref}
\hypersetup{hidelinks}
\usepackage{textcomp}
\usepackage{booktabs}

\def\tsc#1{\csdef{#1}{\textsc{\lowercase{#1}}\xspace}}
\tsc{WGM}
\tsc{QE}
\tsc{EP}
\tsc{PMS}
\tsc{BEC}
\tsc{DE}
\begin{document}
\let\WriteBookmarks\relax
\def\floatpagepagefraction{1}
\def\textpagefraction{.001}
\shorttitle{A WDLoRA-Based Multimodal Generative Framework}
\shortauthors{Xin Zhang et~al.}

\title [mode = title]{A WDLoRA-Based Multimodal Generative Framework for Clinically Guided Corneal Confocal Microscopy Image Synthesis in Diabetic Neuropathy}                      

\author[1]{Xin Zhang}
\ead{x.zhang@mmu.ac.uk}

\author[1]{Liangxiu Han}
\cormark[1]
\ead{l.han@mmu.ac.uk}

\author[1]{Tam Sobeih}
\ead{T.Sobeih@mmu.ac.uk}

\author[1]{Yue Shi}
\ead{y.shi@mmu.ac.uk}

\author[2]{Yalin Zheng}
\ead{yalin.zheng@liverpool.ac.uk}

\author[2]{Uazman Alam}
\ead{uazman.alam@liverpool.ac.uk}

\author[3]{Maryam Ferdousi}
\ead{maryam.ferdousi@manchester.ac.uk}

\author[4]{Rayaz Malik}
\ead{ram2045@qatar-med.cornell.edu}

\affiliation[1]{organization={Department of Computing and Mathematics, Manchester Metropolitan University},
                city={Manchester},
                postcode={M15GD}, 
                country={U.K.}}

\affiliation[2]{organization={Department of Eye and Vision Sciences, University of Liverpool},
                city={Liverpool},
                postcode={L78TX}, 
                country={U.K.}}

\affiliation[3]{organization={Faculty of Biology, Medicine and Health, University of Manchester},
                city={Manchester},
                country={U.K.}}

\affiliation[4]{organization={Department of Medicine, Weill Cornell Medicine-Qatar},
                city={Doha},
                country={Qatar}}

\cortext[cor1]{Corresponding author: Liangxiu Han}

\begin{abstract}
Corneal Confocal Microscopy (CCM) is a sensitive tool for assessing small-fiber damage in Diabetic Peripheral Neuropathy (DPN), yet the development of robust, automated deep learning-based diagnostic models is limited by scarce labelled data and fine-grained variability in corneal nerve morphology. Although Artificial Intelligence (AI)-driven foundation generative models excel at natural image synthesis, they often struggle in medical imaging due to limited domain-specific training, compromising the anatomical fidelity required for clinical analysis. To overcome these limitations, we propose a Weight-Decomposed Low-Rank Adaptation (WDLoRA)-based multimodal generative framework for clinically guided CCM image synthesis. WDLoRA is a parameter-efficient fine-tuning (PEFT) mechanism that decouples magnitude and directional weight updates, enabling foundation generative models to independently learn the orientation (nerve topology) and intensity (stromal contrast) required for medical realism. By jointly conditioning on nerve segmentation masks and disease-specific clinical prompts, the model synthesises anatomically coherent images across the DPN spectrum (Control, T1NoDPN, T1DPN). A comprehensive three-pillar evaluation demonstrates that the proposed framework achieves state-of-the-art visual fidelity (Fr\'echet Inception Distance (FID): 5.18) and structural integrity (Structural Similarity Index Measure (SSIM): 0.630), significantly outperforming GAN and standard diffusion baselines. Crucially, the synthetic images preserve gold-standard clinical biomarkers and are statistically equivalent to real patient data. When used to train automated diagnostic models, the synthetic dataset improves downstream diagnostic accuracy by 2.1\% and segmentation performance by 2.2\%, validating the framework's potential to alleviate data bottlenecks in medical AI.
\end{abstract}

\begin{graphicalabstract}
\includegraphics[width=\linewidth]{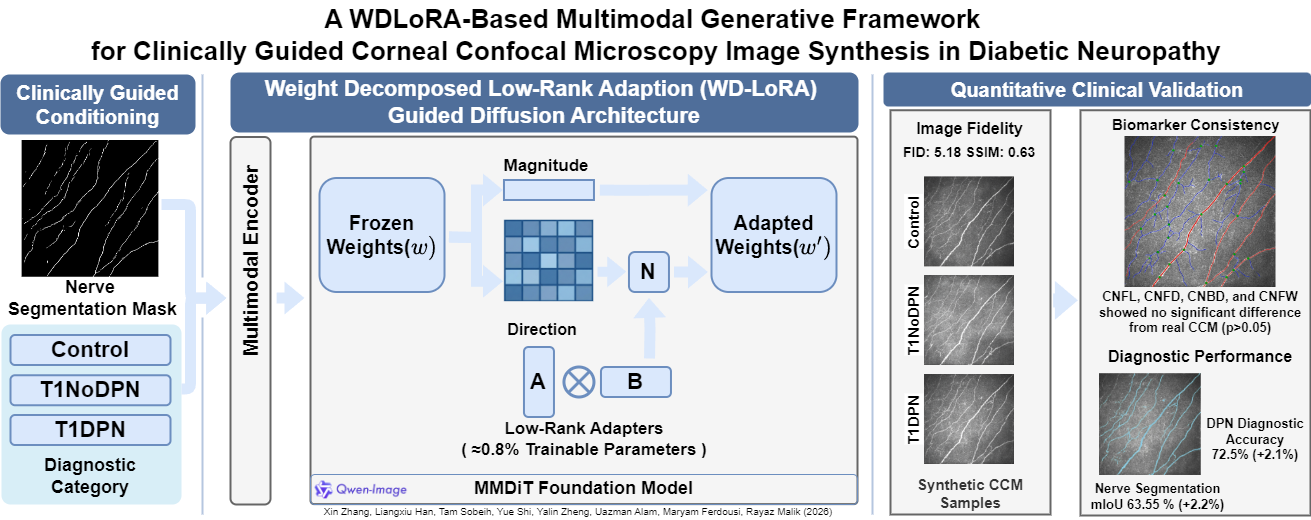}
\end{graphicalabstract}

\begin{highlights}
\item A multimodal generative framework synthesises CCM images using medical priors.
\item WDLoRA enables parameter-efficient fine-tuning of large foundation models.
\item The method achieves state-of-the-art visual fidelity (FID: 5.18) and improves downstream diagnostic and segmentation tasks.
\item An open-access synthetic dataset is released to alleviate annotation bottlenecks.
\end{highlights}


\maketitle

\section{Introduction}
\label{sec:introduction}

Diabetic Peripheral Neuropathy (DPN) is one of the most prevalent microvascular complications of diabetes \cite{Saeedi2019Global}, and early detection is crucial because small-fiber degeneration begins well before clinical symptoms manifest \cite{gad2021corneal}. Corneal Confocal Microscopy (CCM) has emerged as a sensitive, non-invasive modality for quantifying small-fiber pathology, with strong evidence linking CCM biomarkers---corneal nerve fibre length (CNFL), density (CNFD), branch density (CNBD), and fibre width (CNFW)---to DPN severity \cite{petropoulos2020corneal, Petropoulos2021Corneal, Gad2022Corneal, kalteniece2017corneal}. 

Despite this diagnostic utility, the development of reliable deep-learning models for CCM analysis is constrained by intrinsic technical and morphological challenges that make CCM a particularly demanding imaging domain. First, CCM datasets are inherently limited: image acquisition requires expensive instrumentation, precise alignment, trained operators, and consistent corneal contact, resulting in relatively small and highly heterogeneous datasets \cite{Zhou2021review}. Second, CCM images contain ultra-thin, low-contrast nerve fibres---often 2--5 $\mu$m wide---whose visibility is highly sensitive to laser focus, illumination fluctuations, and stromal scattering. These fibres exhibit complex, non-linear geometries, including high tortuosity, variable branching depth, and localised beadings, all of which must be reconstructed with sub-pixel precision. Third, CCM images are affected by modality-specific noise, including speckle noise, non-uniform background illumination, and depth-of-focus artefacts, which complicate both feature extraction and generative modelling \cite{Chen2016automatic}. Moreover, corneal nerve morphology changes gradually across DPN stages, requiring image to represent continuous, fine-scale shifts rather than discrete morphological categories.

Deep Learning have been proposed to automate this analysis \cite{rabah2025deep}, yet current approaches face significant hurdles. Existing models generally fall into two categories: segmentation-based quantification and end-to-end diagnostic classification. While segmentation offers interpretability, it requires pixel-wise annotations that are expensive to generate---creating an ``annotation bottleneck'' \cite{Li2022Context}. Conversely, classification models achieve high diagnostic performance but often lack interpretability, functioning as ``black boxes'' \cite{Dosovitskiy2020Image, Azad2024Advances, Williams2020artificial}. Crucially, both approaches suffer from a universal data bottleneck. Even state-of-the-art models are often trained on modest, single-center datasets \cite{Zhou2021review}. Training high-capacity models on such limited, imbalanced data risks overfitting, resulting in fragile systems that fail to generalize across diverse patient populations or imaging conditions.

GenAI presents a paradigm shift in addressing data scarcity through high-fidelity synthesis. While Generative Adversarial Networks (GANs)\cite{goodfellow2020generative} and Variational Autoencoders (VAEs)\cite{kingma2019introduction} have been explored in medical imaging, they frequently compromise on texture fidelity or suffer from training instability \cite{Sarvamangala2022Convolutional, Azad2024Advances}. Denoising Diffusion Probabilistic Models (DDPMs) \cite{ho2020denoising} have recently set a new benchmark, offering superior mode coverage and stability. Driven by this success, DDPMs have become the base method of modern generative AI, serving for the large foundation generative models such as Hunyuan-DiT \cite{li2024hunyuan}, Stable Diffusion \cite{esser2024scaling,peebles2023scalable}, and Qwen-Image-Edit \cite{wu2025qwen}. These models have achieved outstanding results in natural image generation by learning powerful visual priors from vast datasets. However, due to the lack of domain-specific training data, particularly in ophthalmic domains, generated images may lack fine-grained anatomical fidelity and clinically meaningful pathological variations. To bridge this domain gap, Parameter-Efficient Fine-Tuning (PEFT) techniques, particularly Low-Rank Adaptation (LoRA) \cite{hu2022lora}, have been introduced to adapt large foundation models to specific tasks with minimal computational cost. LoRA adapts foundation models by injecting trainable low-rank matrices while freezing pre-trained weights. However, standard LoRA couples the magnitude and direction of weight updates, limiting the independent control of structural orientation and feature intensity.

To address these limitations, we introduce a multimodal generative framework based on the Qwen-Image-Edit large foundation generative model\cite{wu2025qwen}, incorporating a new Weight-Decomposed Low-Rank Adaptation (WDLoRA) parameter-efficient fine-tuning mechanism. The framework conditions generation on both nerve segmentation masks and clinical prompts (Control, T1NoDPN, T1DPN) to ensure structural and semantic alignment. Crucially, WDLoRA overcomes the limitations of standard LoRA \cite{hu2022lora} by decoupling weight updates into magnitude and direction. This separation enables independent optimization of nerve orientation and intensity—critical for capturing CCM morphology—while preserving pre-trained priors and requiring less than 1\% of parameters to be updated. 
To ensure clinical suitability, we evaluate the framework through a three-pillar protocol encompassing (1) fidelity and diversity metrics, (2) preservation of CCM biomarkers, and (3) downstream diagnostic performance. In summary, this work makes four key contributions:
\begin{itemize}
    \item A WDLoRA-based multimodal generative framework that synthesises CCM images jointly conditioned  on segmentation masks and clinically guided prompts, enabling anatomically coherent and disease-aware image generation.
    \item WDLoRA, a parameter-efficient fine-tuning mechanism that decouples magnitude and directional weight updates, enabling large foundation generative models to specialise to CCM's unique morphological and photometric characteristics while preserving pretrained priors.
    \item A clinically aligned evaluation protocol assessing fidelity/diversity, CCM biomarker preservation, and diagnostic utility, ensuring that the generated images are morphologically faithful and clinically meaningful.
    \item An open-access synthetic dataset featuring clinically aligned diagnostic labels and pixel-wise nerve segmentation masks, released to democratize research and alleviate the annotation bottleneck.
\end{itemize}

\section{Related Work}

Recent GenAI advancements have shifted medical image analysis toward high-fidelity synthesis. This section reviews conditional generative models in healthcare. Additionally, we analyze automated DPN diagnosis via CCM, highlighting the data and annotation bottlenecks that necessitate our proposed framework.

\subsection{Conditional Generative Models for Medical Image Synthesis}

Synthetic Data Generation (SDG) has emerged as a critical solution to the data bottleneck in healthcare, reconciling the demand for large-scale training datasets with stringent privacy regulations like the General Data Protection Regulation (GDPR) and the Health Insurance Portability and Accountability Act (HIPAA) \cite{giuffre2023harnessing, liu2025preserving, junying7synthetic, ezeogu2025synthetic}. By synthesizing high-fidelity, de-identified medical images, SDG facilitates the development of robust AI models for  diseases diagnosis while ensuring patient privacy remains inviolable \cite{khosravi2025exploring, celard2023survey}.  

\begin{figure}
\centering
\includegraphics[width=\linewidth]{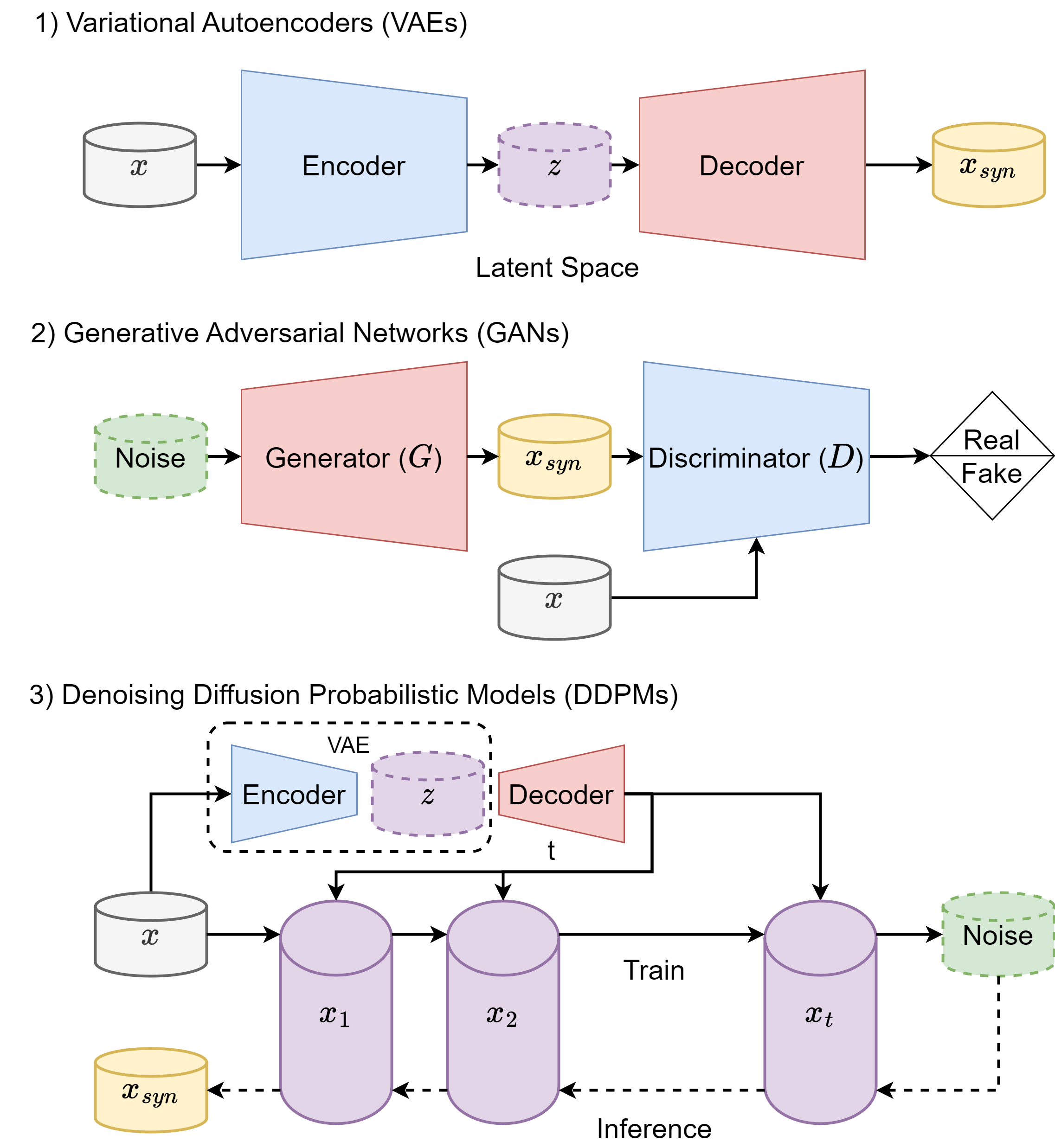}
\caption{Evolution of GenAI Models in Medical Imaging. (1) Variational Autoencoders (VAEs) learn a probabilistic latent space but often produce blurry outputs. (2) Generative Adversarial Networks (GANs) use adversarial training for high fidelity but suffer from mode collapse. (3) Denoising Diffusion Probabilistic Models (DDPMs) iteratively denoise data, offering superior stability and mode coverage, making them the state-of-the-art for medical synthesis.}
\label{fig:generative_models}
\end{figure}

GenAI are a powerful tool that can be broadly categorized into two types: unconditional and conditional. Unconditional models take a random variable as input, allowing for the application of unconditional synthesis. On the other hand, conditional generative models introduce an additional layer of control by incorporating external information or context during the generation process that serve as additional guidance for the model. These can include images, text, semantic maps, class labels, attributes, and signals\cite{ibrahim2025generative}.The conditional GenAI models in medical image analysis can be broadly categorised into VAEs \cite{kingma2019introduction}, GANs \cite{goodfellow2020generative}, DDPMs \cite{ho2020denoising}, flow-based models \cite{kobyzev2020normalizing}, and autoregressive models \cite{van2016conditional}. Among these, DDPMs currently represent the dominant paradigm for high-fidelity and anatomically coherent image synthesis. While alternative approaches such as flow-based and autoregressive models have been explored, their adoption in medical imaging remains comparatively limited. Consequently, this work focuses on VAEs, GANs, and DDPMs, which constitute the most widely used and influential paradigms for medical image synthesis (Fig. \ref{fig:generative_models}).

\textbf{Variational Autoencoders (VAEs).} Emerging in the early 2010s, VAEs represented one of the first successful integrations of deep learning with probabilistic graphical modelling. A VAE is composed of two neural networks: an Encoder and a Decoder. The Encoder ($q_\phi(z|x)$) maps high-dimensional input data ($x$)—such as a CCM image—to a lower-dimensional latent space ($z$). Unlike a standard autoencoder which maps to a fixed vector, a VAE maps the input to a probability distribution (typically a Gaussian), characterised by a mean ($\mu$) and a variance ($\sigma$). The Decoder ($p_\theta(x|z)$) samples a latent vector from this distribution and attempts to reconstruct the original input image\cite{kingma2019introduction}. The training objective is to maximize the Evidence Lower Bound (ELBO):
\begin{equation}
\mathcal{L}_{VAE} = \mathbb{E}_{q_\phi(z|x)}[\log p_\theta(x|z)] - D_{KL}(q_\phi(z|x) || p(z))
\end{equation}
where the first term encourages reconstruction fidelity and the second term (Kullback-Leibler divergence) regularizes the latent space to match the prior $p(z)$. In the medical domain, VAEs have demonstrated versatility across various modalities. For instance, they have been employed to synthesize electronic health records using variational graph autoencoders \cite{nikolentzos2023synthetic} and to generate retinal images for glaucoma assessment \cite{salim2018synthetic}. Advanced architectures like FunSyn-Net have further combined residual VAEs with image-to-image translation to enhance fundus image synthesis \cite{sengupta2020funsyn}. However, VAEs inherently suffer from blurriness due to their Gaussian latent assumption and pixel-wise reconstruction loss, which tends to average out high-frequency details\cite{ibrahim2025generative}. This limitation is critical in CCM generation, where diagnostic value relies on the sharp definition of nerve fibers. Despite these limitations, VAEs remain widely utilized in modern generative models (DDPMs) due to their superiority in data compression. By compressing high-dimensional pixel space into compact latent representations, it significantly reduce the computational load required for training massive foundation models. 

\textbf{Generative Adversarial Networks (GANs).} The introduction of GANs marked a revolution in the visual quality of synthetic images. GANs operate on a game-theoretic premise involving two competing neural networks: a Generator ($G$) and a Discriminator ($D$). The Generator creates synthetic images from random noise, while the Discriminator attempts to distinguish real images from fake ones. These networks engage in a minimax game where the Generator attempts to maximize the Discriminator's error, defined by the value function $V(D,G)$:
\begin{equation}
\begin{split}
\min_G \max_D V(D,G) = \mathbb{E}_{x \sim p_{data}}[\log D(x)] + \\
\mathbb{E}_{z \sim p_{z}}[\log(1 - D(G(z)))]
\end{split}
\end{equation}
This adversarial process forces the generated distribution to align with the real data distribution, producing highly realistic images \cite{goodfellow2020generative}.
In the healthcare domain, GANs have been extensively adopted for medical image synthesis, driven by their ability to generate sharp, high-fidelity textures \cite{islam2024generative}. Beyond standard 2D synthesis, specialized architectures have emerged to address complex modalities; for example, MustGAN employs a multi-stream approach for MRI synthesis \cite{yurt2021mustgan}, while hierarchical amortized GANs have been developed to generate high-resolution 3D medical volumes \cite{sun2022hierarchical}. In ophthalmology, models like CycleGAN and StyleGAN are widely used to synthesize fundus photographs and OCT scans to augment training datasets \cite{zhang2018medical}. However, despite these advancements, GANs face persistent challenges in clinical utility. They are notoriously unstable to train and prone to "mode collapse," where the generator fails to capture the full diversity of the patient population \cite{arora2022generative}. This limitation, coupled with the risk of hallucinating artifacts, complicates their use for robust synthetic data generation in sensitive diagnostic applications.

\textbf{Denoising Diffusion Probabilistic Models (DDPMs).} DDPMs have recently emerged as the state-of-the-art in GenAI, addressing the stability and diversity issues of GANs while surpassing the image quality of VAEs. Inspired by non-equilibrium thermodynamics, diffusion models define a forward process that slowly destroys the structure in a data distribution by adding noise, and a reverse process that learns to restore the structure \cite{ho2020denoising}. The neural network is trained to reverse this process, iteratively denoising the image to recover the original data from pure noise, typically by minimizing the simplified mean squared error between the true noise $\epsilon$ and the predicted noise $\epsilon_\theta$:

\begin{equation}
\mathcal{L}_{simple} = \mathbb{E}_{x_0, \epsilon, t} [\| \epsilon - \epsilon_\theta(\sqrt{\bar{\alpha}_t}x_0 + \sqrt{1-\bar{\alpha}_t}\epsilon, t) \|^2]
\end{equation}
where $x_0$ is the input image, $t$ is the timestep, and $\bar{\alpha}_t$ defines the noise schedule.

Recent studies have demonstrated the efficacy of DDPMs in synthetic data generation across various medical imaging tasks. Khader et al. \cite{khader2023denoising} utilized DDPMs for 3D medical image generation, achieving high-fidelity volumetric synthesis. In the context of segmentation, Saragih et al. \cite{saragih2024using} and Iuliano et al. \cite{iuliano2024denoising} showed that diffusion-generated synthetic data can effectively train robust segmentation models. Within ophthalmology, Alimanov et al. \cite{alimanov2023denoising} and Li et al. \cite{li2025generative} applied DDPMs to retinal images for both generation and segmentation tasks, while Ersari et al. \cite{ersari2025denoising} demonstrated their utility in generating Anterior Segment Optical Coherence Tomography (AS-OCT) images. Diffusion models offer specific advantages for corneal nerve imaging. They rely on a simple mean squared error loss, avoiding the adversarial instability of GANs \cite{ho2020denoising}. Crucially, because they are likelihood-based models, they are mathematically forced to cover the entire data distribution, making them far less likely to drop rare classes (e.g., severe neuropathy cases) compared to GANs \cite{kazerouni2023diffusion}. Furthermore, the iterative nature of the reverse process allows the model to refine details at different scales, resulting in superior fidelity for thin, linear structures like corneal nerves.

\begin{table*}[ht]
\caption{Comparative Analysis of Generative Model Families in Medical Imaging. This comparison highlights the trade-offs between fidelity, diversity, and computational efficiency, justifying the selection of Diffusion Models for high-fidelity CCM synthesis.}
\label{tab:models_comparison}
\centering
\begin{tabular}{p{1.5cm}p{3.5cm}p{3.5cm}p{3.5cm}p{3.5cm}}
\toprule
\textbf{Model Family} & \textbf{Core Principle} & \textbf{Strengths} & \textbf{Limitations} & \textbf{Healthcare Applications} \\
\midrule
\textbf{VAEs}\cite{kingma2019introduction} & Probabilistic encoding into a continuous latent space with reconstruction loss. & Stable training dynamics; excellent sample diversity; smooth latent space suitable for interpolation. & Outputs often suffer from blurriness due to pixel-wise loss; lower perceptual fidelity than GANs. & Latent space analysis; EHRs, Fundus image, 3D medical images\cite{salim2018synthetic, friedrich2024deep,sengupta2020funsyn,nikolentzos2023synthetic}. \\
\addlinespace
\textbf{GANs}\cite{goodfellow2020generative} & Adversarial minimax game between a Generator and a Discriminator. & Creates sharp and realistic images; capable of synthesizing visually convincing textures. & Training instability (non-convergence); prone to "mode collapse" (limited diversity); risk of hallucinating artifacts. & Fundus imaging, OCT synthesis, CT and MRIs\cite{yurt2021mustgan,sun2022hierarchical,bellemo2018generative,tavakkoli2020novel,kumar2022evaluation}.\\
\addlinespace
\textbf{DDPMs}\cite{ho2020denoising} & Iterative denoising process that reverses a gradual forward diffusion chain. & State-of-the-art fidelity and diversity; stable training; comprehensive coverage of the data distribution. & High computational cost; slow inference speeds due to iterative sampling (mitigated by distillations like DDIM). & 3D medical image, Fundus image, OCT image, CCM\cite{khader2023denoising,alimanov2023denoising,ersari2025denoising,li2025generative}. \\
\bottomrule
\end{tabular}
\end{table*}

\textbf{Multimodal Conditional Diffusion Transformer (DiT) generation models and Parameter-Efficient Fine-Tuning (PEFT).} Table \ref{tab:models_comparison} summarizes the trade-offs between fidelity, diversity, and computational efficiency across these generative paradigms. DDPMs offer a superior balance of fidelity and diversity compared to VAEs and GANs, propelling them to the forefront of modern GenAI. They serve as the backbone for large-scale Diffusion Transformer (DiT) models, such as Hunyuan-DiT (1.5B–13B parameters) \cite{li2024hunyuan}, Stable Diffusion 3 (up to 8B parameters) \cite{esser2024scaling,peebles2023scalable}, and Qwen-Image-Edit (20B parameters) \cite{wu2025qwen}. Among these, Qwen-Image-Edit is particularly notable for its integration of the Qwen2.5-VL vision-language model with a Multimodal Diffusion Transformer (MMDiT) backbone. This architecture enables it to process visual and textual tokens as a unified sequence, facilitating precise instruction following and fine-grained local editing—capabilities that are essential for medical image synthesis where anatomical fidelity is paramount. These models leverage vast internet-scale datasets to learn powerful visual priors, enabling photorealistic synthesis and complex semantic editing. However, they often struggle with domain-specific medical tasks due to the scarcity of specialized training data. For instance, synthetic CCM images must rigorously adhere to biological constraints, accurately reproducing the tortuosity of sub-basal nerve fibers while independently managing photometric variances caused by imaging artifacts and corneal opacity.

Adapting these large foundation generative models to data-scarce medical domains requires PEFT to bridge the domain gap without overfitting. Standard PEFT techniques, such as Low-Rank Adaptation (LoRA) \cite{hu2022lora} and its variants \cite{zhang2023adalora,valipour2023dylora}, inject low-rank adapter matrices into Transformer blocks while freezing most pre-trained weights. A critical limitation of standard LoRA in this context is that it inherently couples the magnitude and direction of feature updates. Recently, Weight-Decomposed Low-Rank Adaptation (DoRA) \cite{liu2024dora} was introduced to decouple these components in large language models, demonstrating that directional updates dominate learning capacity while magnitude governs scaling. However, in sensitive medical domains like CCM imaging, pathological features such as nerve fiber orientation (structural direction) and signal intensity (photometric magnitude) vary independently and require explicitly decoupled control tailored to multimodal diffusion architectures to ensure anatomical realism. To our knowledge, no existing CCM synthesis framework integrates spatial anatomical conditioning with such decoupled adaptation. This gap motivates our proposed Weight-Decomposed Low-Rank Adaptation (WDLoRA) strategy for MMDiT models, which enables independent optimization of structural and photometric features within a unified, clinically guided generative pipeline.

\subsection{Diabetic Peripheral Neuropathy Diagnosis using CCM}

DPN is a common complication of diabetes, characterised by progressive degeneration of small peripheral nerve fibres. Conventional diagnostic methods, such as nerve conduction studies, primarily assess large fibres and are therefore insensitive to early neuropathic changes. CCM has emerged as a non-invasive imaging modality that enables in vivo visualisation of the corneal sub-basal nerve plexus and provides a sensitive surrogate biomarker for DPN.

Traditional diagnostic methods for DPN, such as nerve conduction studies, primarily assess large nerve fibers and often fail to detect early small-fiber pathology. CCM has emerged as a powerful, non-invasive imaging modality that provides a ``window'' into the peripheral nervous system, allowing for the in vivo visualisation of the sub-basal nerve plexus. The degeneration of these corneal nerves has been shown to correlate strongly with the severity of systemic neuropathy, establishing CCM as a robust surrogate biomarker for DPN \cite{Iqbal2018Diabetic, Burgess2021Early}.

The identification of DPN via CCM relies on the quantification of specific morphological parameters that reflect nerve health. International consensus has defined several key metrics. These parameters provide a comprehensive profile of corneal nerve pathology, capturing both the degenerative (loss of length and density) and regenerative (branching and width changes) aspects of the disease process.
\begin{itemize}
    \item \textbf{Corneal Nerve Fiber Length (CNFL):} Defined as the total length of all nerve fibers and branches per square millimeter (mm/mm$^2$). CNFL is widely considered the most sensitive and reproducible single metric for early DPN diagnosis, with reductions consistently observed before the onset of clinical symptoms \cite{Petropoulos2021Corneal}.
    \item \textbf{Corneal Nerve Fiber Density (CNFD):} The total number of major nerve trunks per square millimeter (no./mm$^2$). A decrease in CNFD is a specific indicator of established neuropathy \cite{kalteniece2017corneal}.
    \item \textbf{Corneal Nerve Branch Density (CNBD):} The number of branch points on major nerve trunks per square millimeter (no./mm$^2$). This metric reflects the regenerative capacity of the nerve plexus and is often the first to decline in early-stage disease \cite{kalteniece2017corneal}.
    \item \textbf{Corneal Nerve Fiber Width (CNFW):} While less commonly reported than length or density, nerve width (or diameter) offers complementary information regarding axonal health. Measured in micrometers ($\mu$m), changes in width may reflect early axonal swelling due to osmotic stress or subsequent atrophy, although measurement variability has historically limited its widespread clinical use \cite{Gad2022Corneal}.
\end{itemize}

However, manual quantification of corneal nerve morphology is labour-intensive and subject to inter-observer variability, motivating the adoption of deep learning for automated CCM analysis. Early approaches focused on semantic segmentation, utilizing architectures like U-Net and its variants (e.g., CS-Net, NerveFormer) to automate the tracing of nerve fibers \cite{ronneberger2015unet, Li2022Context}. These models aim to replicate human annotation, enabling the automated calculation of CNFL and CNFD. More recently, ``end-to-end'' classification models (e.g., ResNet, Vision Transformers) have been developed to directly map CCM images to diagnostic labels (DPN+ vs. DPN-), achieving high sensitivity by learning global image features \cite{Williams2020artificial, Meng2023Artificial}.

These limitations motivate the use of synthetic data generation to augment real CCM datasets. However, research specifically targeting CCM image synthesis—particularly for DPN diagnosis—remains scarce, and existing works often struggle with generalization due to small, noisy, and heterogeneous datasets\cite{qiao2025connect,ibrahim2025generative}. Moreover, clinically meaningful synthesis requires more than visual realism; it necessitates explicit anatomical and pathological control to preserve nerve topology, continuity, and disease-specific morphological variation. To bridge these gaps, we introduce a multimodal conditional generative framework based on a large DiT model. By conditioning synthesis on nerve segmentation masks and clinical prompts, our approach generates high-fidelity, disease-specific CCM images that accurately reflect DPN progression, thereby enhancing the robustness of downstream diagnostic models.

\section{Methods}

\subsection{Overview of the Proposed Conditional Multimodal Generative Framework}

In this work, we propose a clinically conditional multimodal diffusion framework for CCM images that explicitly model disease-specific nerve morphology associated with DPN. Rather than relying on unconstrained text-to-image generation, the proposed method introduces joint spatial--semantic control through the integration of anatomical priors, clinically structured prompts, and a parameter-efficient adaptation strategy. The framework is built upon the Qwen-Image-Edit multimodal diffusion foundation model \cite{wu2025qwen}, but substantially extends it in three key directions:
\begin{enumerate}
    \item Explicit anatomical conditioning via nerve segmentation masks;
    \item Clinically stratified semantic control aligned with diagnostic cohorts; and
    \item Weight-Decomposed Low-Rank Adaptation (WDLoRA) to enable stable, fine-grained medical-domain specialisation.
\end{enumerate}

Given a nerve segmentation mask and a clinically targeted prompt, the model synthesises high-fidelity CCM images that preserve nerve continuity, spatial topology, and disease-relevant morphological variations. We define three clinically meaningful conditioning categories reflecting progressive neuropathic severity: \textit{Control}, \textit{Type 1 Diabetes without DPN (T1NoDPN)}, and \textit{Type 1 Diabetes with DPN (T1DPN)}. These categories are encoded directly into the generation process rather than inferred implicitly from data.

\begin{figure}
\centering
\includegraphics[width=\linewidth]{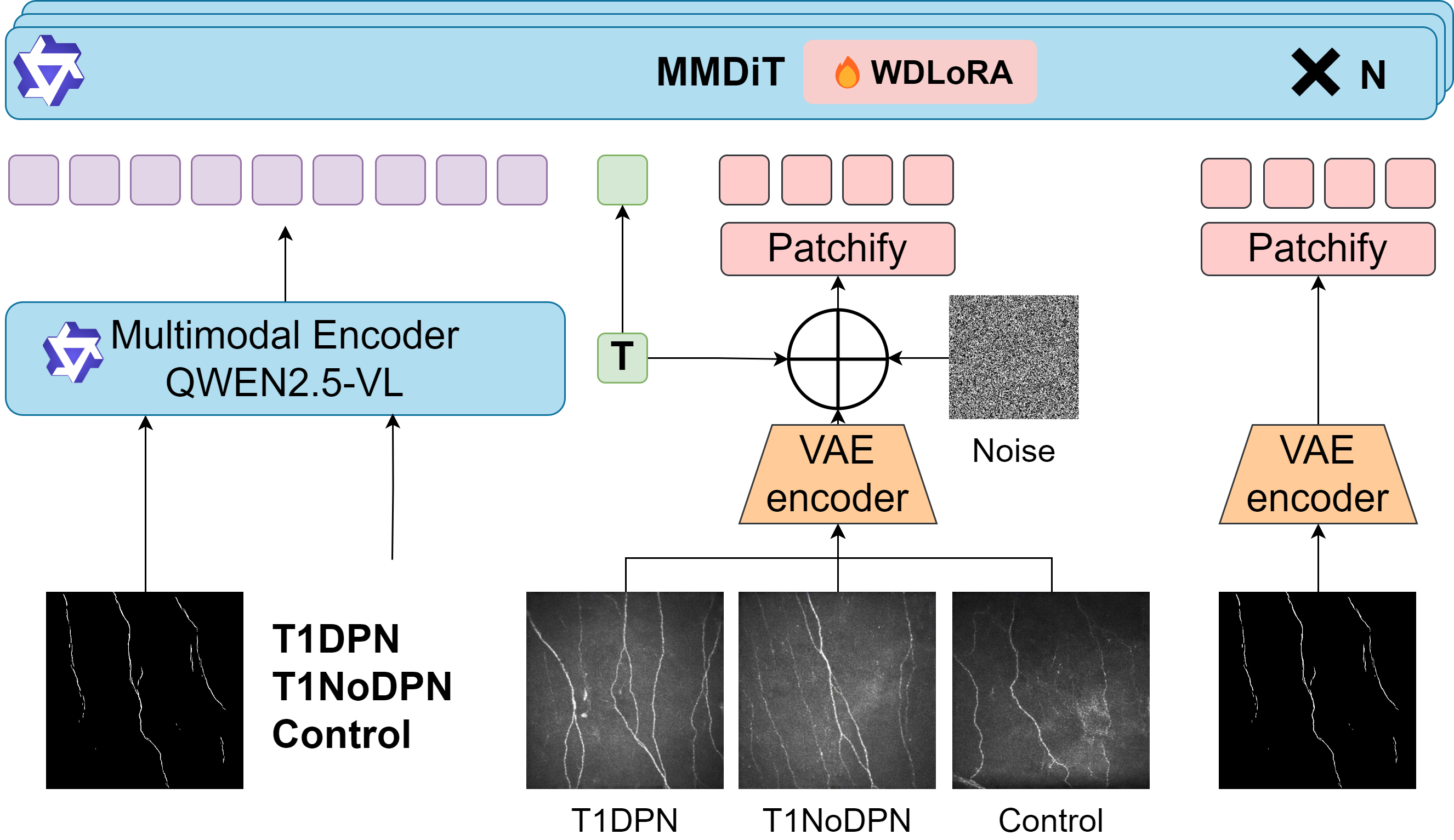} % Replace with actual architecture diagram
\caption{Overview of the proposed generative framework. Building on Qwen-Image-Edit \cite{wu2025qwen}, the pipeline uses the Qwen2.5-VL encoder to extract semantic features from multimodal inputs (nerve segmentation masks and clinical text prompts). These features condition the Multimodal Diffusion Transformer (MMDiT) backbone. Weight-Decomposed Low-Rank Adaptation (WDLoRA) is used to efficiently fine-tune the MMDiT blocks on CCM data, enabling high-fidelity synthesis while preserving pre-trained knowledge.}
\label{fig:qwen_arch}
\end{figure}

Figure \ref{fig:qwen_arch} illustrates the proposed framework. The generation pipeline operates by encoding multimodal inputs---specifically nerve segmentation masks, clinically structured text prompts, and optional reference images---into a shared semantic space using the Qwen2.5-VL encoder. These representations condition a latent diffusion process governed by a Multimodal Diffusion Transformer (MMDiT), which iteratively denoises latent variables. This design explicitly targets three CCM-specific requirements:
\begin{enumerate}
    \item Structural fidelity: preservation of corneal nerve fibre layout, branching continuity, and global topology.
    \item Photometric realism: accurate modelling of stromal background texture and imaging characteristics.
    \item Pathology-aware variation: controlled modulation of nerve density, tortuosity, and fragmentation across disease stages (Control $\to$ T1NoDPN $\to$ T1DPN).
\end{enumerate}

By enforcing these explicit clinical and anatomical priors throughout the denoising trajectory, the framework ensures that generated images remain both visually realistic and diagnostically meaningful, unlike generic multimodal generation pipelines.

\subsection{Multimodal Conditional Image Generation Foundation Model}

We adopt Qwen-Image-Edit, a state-of-the-art foundation model for multimodal conditional image generation, as the backbone due to its strong joint text--image representation capability and transformer-based diffusion design. Departing from conventional U-Net-based diffusion approaches, it uses a Multimodal Diffusion Transformer (MMDiT) backbone. By processing visual and textual tokens as a unified sequence, the model applies self-attention across both modalities, enabling tighter alignment between instructions and visual content. Moreover, transformer backbones scale more effectively with data and parameters than CNN-based U-Nets, supporting the modelling of subtle, fine-grained anatomical variations. Under multimodal conditional control, this architecture can preserve complex semantic relationships and anatomical details that are critical for capturing DPN-related morphological changes. The model integrates Qwen2.5-VL for semantic guidance from prompts and images, while employing a VAE encoder to maintain visual fidelity, thereby supporting both high-level semantic consistency and low-level structural control.

\subsubsection{Multimodal Diffusion Transformer (MMDiT)}

\begin{figure}
\centering
\includegraphics[width=\linewidth]{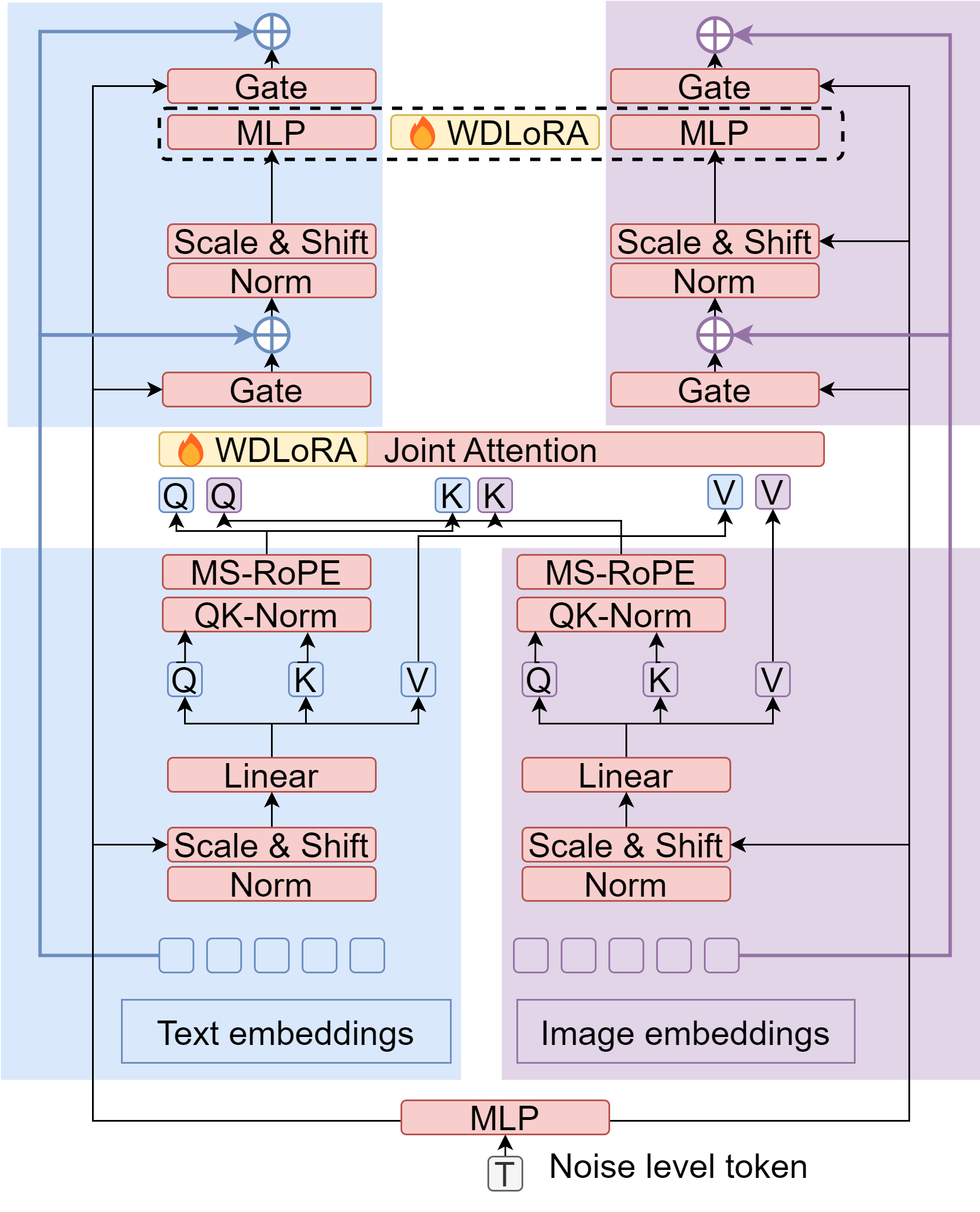}
\caption{Architecture of the Multimodal Diffusion Transformer (MMDiT). The model processes concatenated image and text tokens through a unified self-attention mechanism, enabling bidirectional cross-attention for precise semantic alignment.}
\label{fig:mmdit_arch}
\end{figure}

The MMDiT processes visual and textual representations as a single unified token sequence, enabling deep bidirectional conditioning between modalities. Unlike traditional approaches that rely on separate cross-attention layers, the model jointly embeds and attends to image tokens (derived from latent representations and segmentation masks) and text tokens (encoding diagnostic semantics) via self-attention. This allows semantic instructions to directly influence spatial structure throughout the denoising process (Fig. \ref{fig:mmdit_arch}). Formally, queries ($Q$), keys ($K$), and values ($V$) from both modalities are concatenated into shared sequences:

\begin{equation}
\begin{aligned}
Q &= [Q_{img} \oplus Q_{txt}] \\
K &= [K_{img} \oplus K_{txt}] \\
V &= [V_{img} \oplus V_{txt}]
\end{aligned}
\label{eq:attention_concat}
\end{equation}

where $\oplus$ denotes concatenation. This all-to-all attention mechanism allows the model to capture long-range dependencies, which is critical for maintaining the structural continuity of long corneal nerve fibers without fragmentation. 
To effectively model the 2D spatial structure of the image within this 1D sequence, MMDiT incorporates Multimodal Scalable Rotary Positional Embeddings (MSRoPE). In this approach, text inputs are treated as pseudo-2D tensors with identical position IDs applied across both dimensions, conceptually placing text tokens along the diagonal of the image spatial grid. This design enables MSRoPE to leverage resolution scaling advantages for the image modality while maintaining functional equivalence to 1D-RoPE for text, thereby eliminating the need for heuristic text positional encoding strategies. Consequently, the model generalizes robustly to varying aspect ratios and resolutions, a critical feature for processing CCM images with diverse fields of view.

\subsubsection{Multimodal Encoding and Latent Diffusion Space}

Multimodal conditioning signals are encoded using the Qwen2.5-VL encoder, which produces a unified semantic representation from heterogeneous inputs, including clinically structured text prompts, nerve segmentation masks, and optional reference CCM images. Rather than serving as auxiliary guidance, these representations are injected directly into the diffusion process and modulate the denoising dynamics at every timestep, enforcing persistent alignment between anatomical structure and diagnostic semantics.

The segmentation mask provides an explicit spatial prior, anchoring the global topology and continuity of the corneal nerve plexus, while the text prompt encodes disease-stage--specific morphological characteristics (e.g., reduced nerve density or increased tortuosity). By embedding both signals within a shared semantic space, the model jointly reasons about where structures should appear and how they should morphologically manifest under different pathological conditions.

To ensure computational tractability and training stability, diffusion is performed in a compressed latent space. A pre-trained VAE maps high-resolution CCM images into low-dimensional latent representations, substantially reducing memory and compute requirements. Operating in latent space allows the diffusion transformer to prioritise high-level structural organisation and semantic consistency, rather than expending capacity on low-level pixel reconstruction.

Clinical control is further reinforced through system-level prompt stratification. Separate prompt templates are defined for the three diagnostic cohorts---Control, T1NoDPN, and T1DPN---ensuring that disease-stage semantics are explicitly encoded as conditioning variables throughout training and inference. This design prevents pathological characteristics from emerging implicitly via dataset bias and instead enforces controlled, interpretable variation aligned with clinical taxonomy. 

\subsection{Weight-Decomposed Low-Rank Adaptation (WDLoRA)}

To adapt such models to the highly specialized CCM, we introduce a parameter-efficient fine tuning method WDLoRA, building upon the theoretical foundations of Weight-Decomposed Low-Rank Adaptation (DoRA) \cite{liu2024dora}.

Standard LoRA approximates the weight update $\Delta W$ for a pre-trained matrix $W_0 \in \mathbb{R}^{d \times k}$ via two low-rank matrices $B \in \mathbb{R}^{d \times r}$ and $A \in \mathbb{R}^{r \times k}$ (where $r \ll \min(d, k)$), such that $W' = W_0 + BA$. While computationally efficient, this formulation inherently couples the magnitude and directional updates of the weight vectors. This rigid coupling limits the model's ability to make the nuanced, fine-grained adjustments necessary to capture subtle morphological features like corneal nerve tortuosity.

WDLoRA addresses this limitation by decoupling the optimization of feature orientation and scaling, extending DoRA's concepts to the multimodal diffusion setting. We re-parameterise the weight matrix $W$ into a magnitude vector $m \in \mathbb{R}^d$ and a directional matrix $V \in \mathbb{R}^{d \times k}$:

\begin{equation}
W = m \odot \frac{V}{\|V\|_c}
\end{equation}

where $\|V\|_c$ denotes the column-wise Euclidean norm, and $\odot$ represents element-wise multiplication (broadcasting $m$ across columns).

During fine-tuning, the pre-trained directional component $V$ is frozen and adapted via a low-rank update ($BA$), while the magnitude vector $m$ is trained as a standalone parameter. The update rule is formulated as:

\begin{equation}
W' = (m + \Delta m) \odot \frac{V + BA}{\|V + BA\|_c}
\end{equation}

\begin{figure*}
\centering
\includegraphics[width=\linewidth]{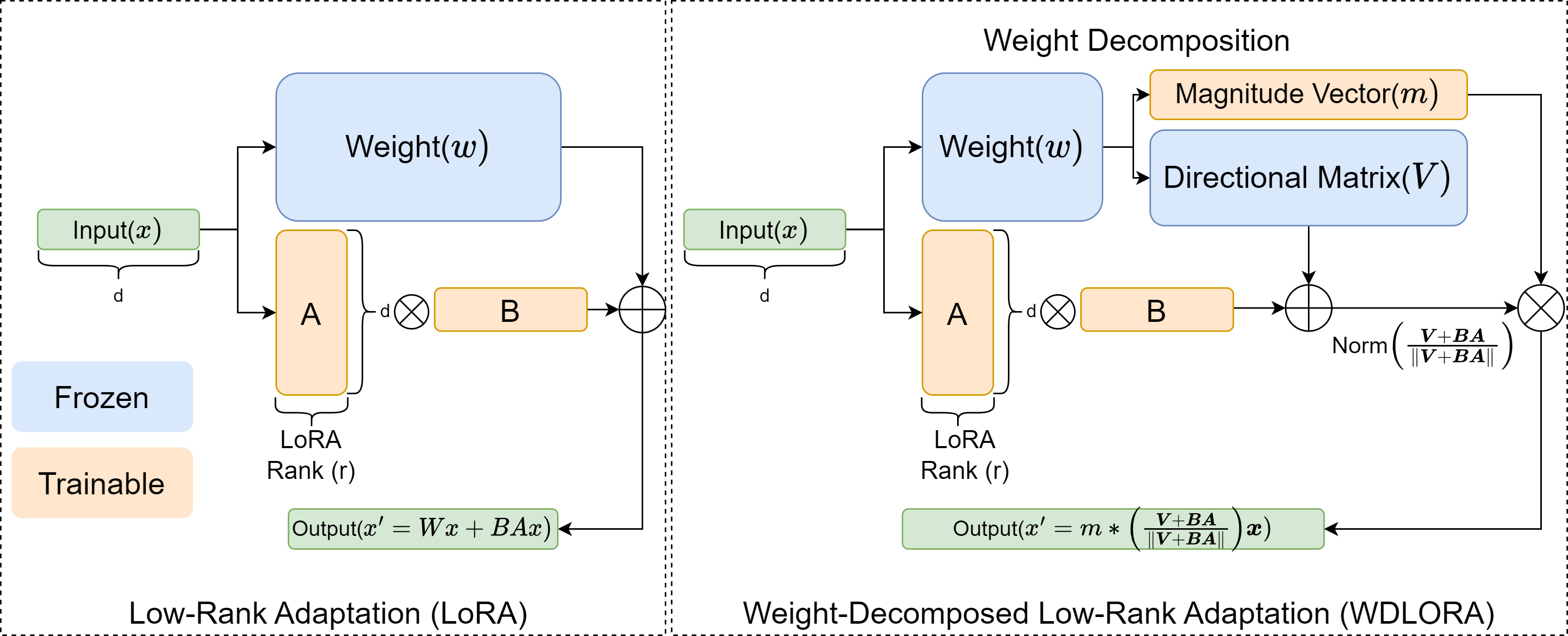} % Placeholder for WDLoRA diagram
\caption{Schematic of Weight-Decomposed Low-Rank Adaptation (WDLoRA). The mechanism decomposes weights into magnitude and direction, applying low-rank updates only to the directional component. This structure is applied to the Attention and MLP layers of the MMDiT backbone.}
\label{fig:WDLoRA_diagram}
\end{figure*}

This decomposition allows the model to independently refine the \textit{orientation} of features (critical for defining nerve structure) and their \textit{amplitude} (essential for contrast and texture intensity). Fig.~\ref{fig:WDLoRA_diagram} illustrates this mechanism.

Implementation-wise, WDLoRA is applied selectively to the attention and feed-forward layers of the MMDiT. Specifically, we replace the conventional linear layers in the dimension mapping modules and selectively apply WDLoRA to the query, key, value, and output projection layers within the transformer blocks, while the remainder of the foundation model remains frozen. This design yields several key advantages:
\begin{itemize}
    \item \textbf{Fine-grained morphological control:} Subtle pathological variations are captured without disrupting global semantic alignment.
    \item \textbf{Stable optimisation:} Decoupled updates reduce gradient interference during diffusion training.
    \item \textbf{Parameter efficiency:} By injecting these lightweight adapters, we reduce the number of trainable parameters by over 99\% compared to full fine-tuning, enabling adaptation with limited CCM data.
    \item \textbf{Preservation of pretrained priors:} Natural image and language knowledge encoded in the foundation model is retained.
\end{itemize}

Overall, WDLoRA allows the DiT model to progressively refine nerve morphology across disease stages, supporting controlled transitions from healthy plexus organisation (Control) to reduced density and increased tortuosity (T1DPN), while maintaining anatomical plausibility.

\section{Experiments and Evaluaion}

\subsection{Dataset and Preprocessing}
The dataset utilized in this study was provided by the Early Neuropathy Assessment (ENA) group at the University of Manchester, UK. It comprises in vivo CCM images of the sub-basal nerve plexus acquired from a cohort of 318 individuals, including healthy volunteers and patients with Type-1 diabetes.

\subsubsection{Image Acquisition}
Images were captured using the Rostock Corneal Module of the Heidelberg Retina Tomograph III (HRT III-RCM, Heidelberg Engineering, Germany). Following an internationally recognized protocol \cite{kalteniece2017corneal}, images were acquired at a resolution of $384 \times 384$ pixels, covering a field of view of $400 \times 400$ $\mu$m. To ensure high data quality, 4 to 8 non-overlapping images with optimal contrast and focus were selected per participant. All images were exported in BMP format.
The study population was categorised into three distinct clinical groups based on established diagnostic criteria:
\begin{itemize}
    \item \textbf{Healthy Volunteers (Control):} 112 participants (35.2\%) with no history of diabetes or neuropathy.
    \item \textbf{Type 1 Diabetes without Neuropathy (T1noDPN):} 98 patients (30.8\%) with a diabetes duration of $\geq$ 5 years but no clinical signs of neuropathy.
    \item \textbf{Type 1 Diabetes with Neuropathy (T1DPN):} 108 patients (34.0\%) with confirmed DPN symptoms.
\end{itemize}

Detailed demographic and clinical characteristics of the study participants are summarized in Table \ref{tab:demographics}.

\begin{table}[h]
\centering
\caption{Demographic and clinical characteristics of the study participants. Values are presented as Mean $\pm$ Standard Deviation (SD) or Count.}
\label{tab:demographics}
\resizebox{0.5\textwidth}{!}{%
\begin{tabular}{cccc}
\hline
Variable & Control & with DPN (PN+) & without DPN (PN-) \\ \hline
Duration of diabetes (years) & 0 $\pm$ 0 & 45.54 $\pm$ 11.96 & 23.5 $\pm$ 13.74 \\
Age & 39.64 $\pm$ 14.15 & 60.79 $\pm$ 11.40 & 41.85 $\pm$ 14.46 \\
Gender (Female/Male) & 56.4\%/43.6\% & 52.8\%/47.2\% & 44.6\%/55.4\% \\
BMI ($kg/m^2$) & 25.64 $\pm$ 4.74 & 27.55 $\pm$ 3.90 & 26.22 $\pm$ 4.56 \\
HbA1c & 5.56 $\pm$ 0.34 & 8.27 $\pm$ 1.34 & 8.34 $\pm$ 1.45 \\
IFCC (mmol/mol) & 36.34 $\pm$ 5.51 & 66.60 $\pm$ 14.25 & 67.60 $\pm$ 15.79 \\ \hline
\end{tabular}%
}
\end{table}

\subsubsection{Data Preparation for Generative Modeling}
\begin{figure}
\centering
\includegraphics[width=\linewidth]{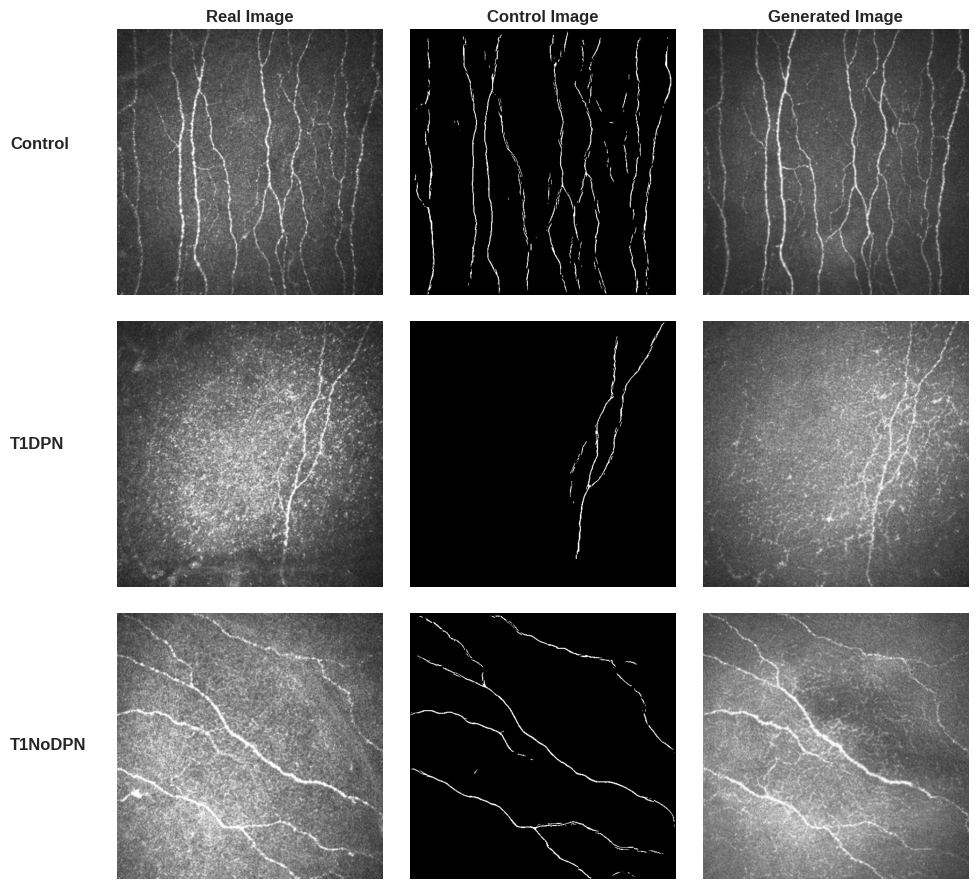}
\caption{visualisation of the dataset used in this study. The figure displays the diagnostic class labels and the corresponding expert annotations (nerve segmentation masks), which serve as the control images for the generative model.}
\label{fig:com}
\end{figure}

To enable the training of our conditional generative model, we prepared a paired dataset consisting of raw CCM images, diagnostic class labels, and corresponding nerve segmentation masks, as shown in Fig. \ref{fig:com}. Expert annotators manually traced the corneal nerve fibers to generate precise pixel-wise binary segmentation masks. These masks serve a dual purpose: they act as ground truth for the downstream segmentation evaluation and as structural conditioning inputs for the generative model. During training, the model receives the segmentation mask (representing the nerve morphology) and the diagnostic class label (representing the pathological state) to learn the mapping to the realistic texture and appearance of the CCM image.

\subsection{Three-pillar Evaluation Framework}

To properly validate the proposed generative framework, we established a comprehensive ``Three-pillar Evaluation Framework.'' This protocol assesses the synthetic data from three complementary dimensions: technical visual fidelity, clinical morphological validity, and practical utility in downstream medical tasks.

\subsubsection{Pillar 1: Technical Visual Fidelity and Diversity Analysis}
The first pillar focuses on evaluating the generation quality from a computer vision perspective. We employ quantitative metrics grounded in statistical feature analysis to evaluate the distributional alignment between real ($r$) and synthetic ($g$) domains.

\textbf{Fréchet Inception Distance (FID), Peak Signal-to-Noise Ratio (PSNR) and Structural Similarity Index Measure (SSIM):} We adopt FID as the primary metric to quantify the distance between the feature distributions of real and synthetic images. A lower FID indicates that the generated images are more realistic and possess a diversity distribution similar to the real dataset. PSNR quantifies reconstruction quality by measuring the ratio between the maximum signal power and corrupting noise. SSIM assesses the preservation of high-frequency corneal structures based on luminance, contrast, and structure. High scores in these metrics confirm that the generated images preserve fine-grained textural details of the corneal stroma and nerve fibers.

\textbf{Diversity Analysis:} To ensure the model generates diverse pathological examples rather than memorizing training data, we employ Intra-class Diversity (measuring variation within samples of the same class) and Inter-class Separability (evaluating discriminability between different classes).

\subsubsection{Pillar 2: Clinical Morphological Validity}
\begin{figure}
\centering
\includegraphics[width=\linewidth]{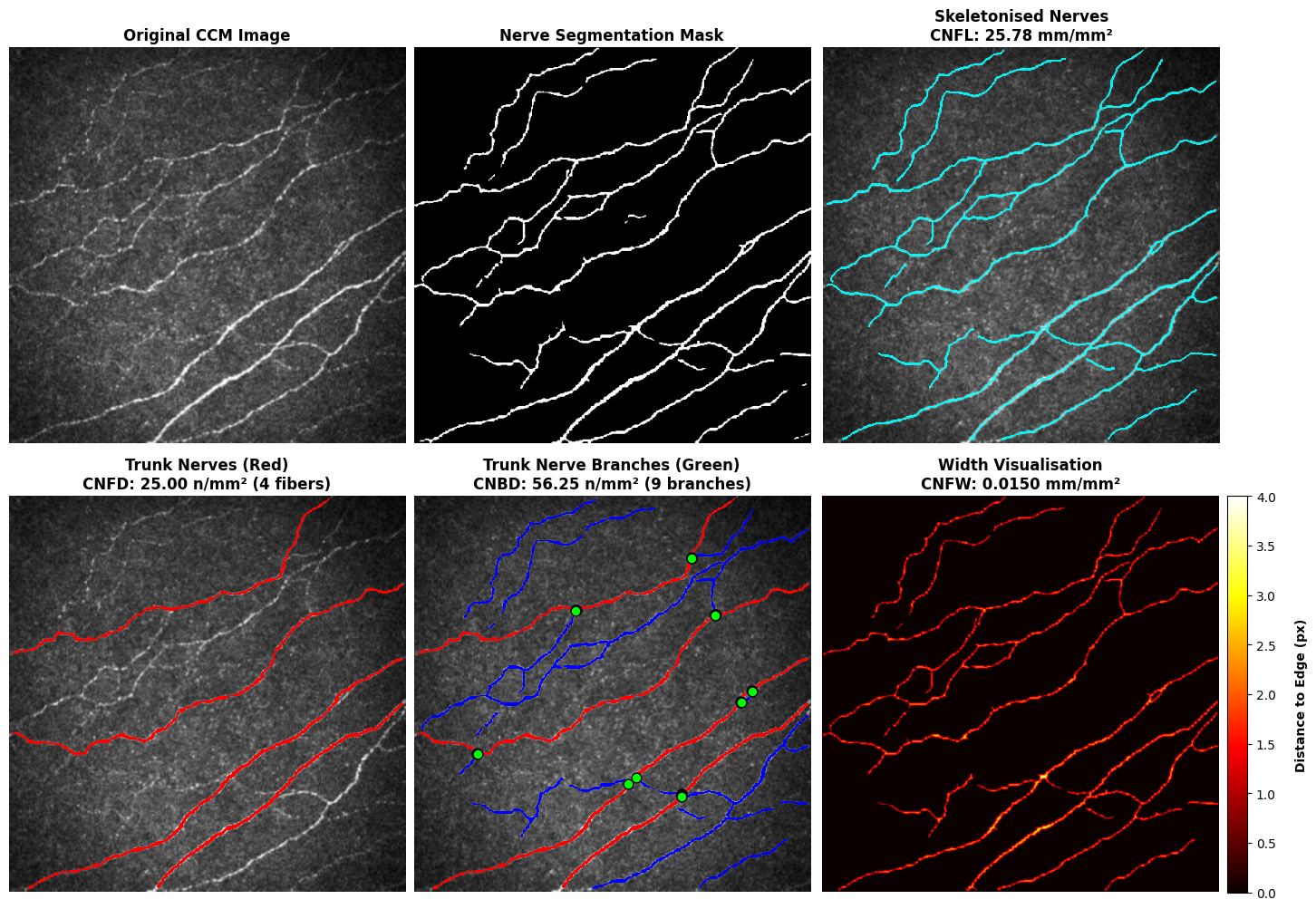}
\caption{Visualization of key Corneal Confocal Microscopy (CCM) biomarkers used for DPN diagnosis. Top row: Original CCM image, Nerve Segmentation Mask, and Skeletonized Nerves (used for CNFL). Bottom row: Main Trunks (CNFD), Branch Points (CNBD), and Nerve Width Visualization (CNFW). These morphological parameters serve as the gold standard for quantifying small-fiber pathology.}
\label{fig:ccm}
\end{figure}

Visual realism alone is insufficient for medical applications; the synthetic data must be biologically accurate (Fig. \ref{fig:ccm}). We validate the generated images against ``gold standard'' morphological biomarkers used in the clinical diagnosis of DPN: CNFL, CNFD, CNBD, and CNFW. To extract these metrics uniformly, we employ an automated morphological extraction pipeline. Generated images are first segmented using the baseline HMSViT model. The resulting masks are then skeletonized using the Zhang-Suen thinning algorithm to extract single-pixel-wide nerve representations, enabling the identification of main trunks and branch points. Using the imaging calibration parameters (a $400 \times 400$ $\mu$m field of view natively captured over $384 \times 384$ pixels, yielding approximately $1.04$ $\mu$m/pixel), pixel lengths are converted into standard physical units ($\mu$m). By ensuring statistical alignment of these extracted metrics against the corresponding clinical ground truth, we verify that the synthetic images faithfully replicate the actual underlying microscopic pathology.

\subsubsection{Pillar 3: Downstream Task Utility}
The final validation lies in the practical utility of synthetic data for medical AI development. We evaluate this by applying the data to two critical downstream tasks: \textbf{DPN Diagnosis} (classification) and \textbf{Nerve Segmentation}. We assess whether augmenting real-world data with synthetic samples improves model performance, thereby serving as a proxy for the data's semantic and structural quality.

\subsection{Experiment design}

Based on our three-pillar evaluation framework, we designed three experiments to validate the fidelity, clinical utility, and comparative performance of the proposed framework.

\subsubsection{Experiment 1: Image Generation Quality Evaluation}

This experiment evaluates the visual fidelity, clinical accuracy, and diversity of the generated CCM images. We systematically compare the proposed WDLoRA against the traditional LoRA method across three rank configurations (Rank 8, 16, and 32), as detailed in Table \ref{tab:model_params}. The assessment is conducted from three complementary perspectives: (1) \textbf{General Image Quality}, using standard computer vision metrics including FID, PSNR, and SSIM. FID is calculated as:
\begin{equation}
FID = \|\mu_{r} - \mu_{g}\|^{2} + Tr\left(\Sigma_{r} + \Sigma_{g} - 2(\Sigma_{r}\Sigma_{g})^{1/2}\right)
\end{equation}
where $(\mu, \Sigma)$ denote the mean and covariance of features extracted via Inception-v3. PSNR measures the reconstruction quality and is defined as:
\begin{equation}
PSNR = 10 \cdot \log_{10}\left(\frac{MAX_I^2}{MSE}\right)
\end{equation}
where $MAX_I$ is the maximum pixel value and $MSE$ is the mean squared error between the real and generated images. SSIM evaluates structural similarity:
\begin{equation}
SSIM(x, y) = \frac{(2\mu_x\mu_y + C_1)(2\sigma_{xy} + C_2)}{(\mu_x^2 + \mu_y^2 + C_1)(\sigma_x^2 + \sigma_y^2 + C_2)}
\end{equation}
where $\mu_x, \mu_y$ are local means, $\sigma_x, \sigma_y$ are standard deviations, and $\sigma_{xy}$ is the cross-covariance.

(2) \textbf{Clinical Morphological Fidelity}, by quantifying key corneal nerve parameters—Corneal Nerve Fiber Length (CNFL), Density (CNFD), Branch Density (CNBD), and Width (CNFW); and (3) Diversity Analysis, employing t-SNE visualisation and quantitative metrics (Intra-class Diversity and Inter-class Separability) to ensure the model captures the full spectrum of pathological variations. We compare these metrics between synthetic images and real ground truth data across three pathological groups (Control, T1NoDPN, T1DPN) to determine the optimal configuration for high-fidelity synthesis and validate the superior adaptation capabilities of WDLoRA.

\begin{table}[h]
\centering
\caption{Comparison of Trainable Parameters. A comparative analysis of parameter efficiency between Standard LoRA and the proposed WDLoRA mechanism across varying rank configurations (8, 16, 32). WDLoRA introduces negligible computational overhead (<0.01\% parameter increase) while significantly altering the adaptation dynamics.}
\label{tab:model_params}
\begin{tabular}{cccc}
\toprule
\textbf{Rank} & \textbf{LoRA} & \textbf{WDLoRA} & \textbf{Params \% of total} \\
\midrule
8 & 11.80M & 11.81M & 0.11\% \\
16 & 23.59M & 23.61M & 0.22\% \\
32 & 47.19M & 47.21M & 0.44\% \\
\bottomrule
\end{tabular}
\end{table}

\subsubsection{Experiment 2: Medical Task Performance Evaluation}

This experiment aims to demonstrate the practical utility of our synthetic data by quantifying its impact on two critical medical tasks: \textbf{DPN Diagnosis} (classification) and \textbf{Nerve Segmentation}. We evaluate the performance benefits of generative data augmentation across three training regimes: Group A (Baseline), trained exclusively on the limited real-world dataset; Group B (Hybrid), trained on a hybrid dataset composed of real images augmented with our synthetic data; and Group C (Synthetic-Only), trained exclusively on the synthetic dataset.

For both the classification and segmentation tasks, we employ the state-of-the-art HMSViT \cite{zhang2025hmsvit} model. Specifically, for the classification task, we train HMSViT to categorize CCM images into three diagnostic groups: Control, T1NoDPN, and T1DPN. For the segmentation task, we utilize HMSViT to delineate corneal nerve fibers. All models are evaluated on a held-out test set of real patient images using standard metrics, including Accuracy for diagnosis and mean Intersection over Union (mIoU) for segmentation.

\subsubsection{Experiment 3: Comparative Analysis of Generative Models}
To evaluate the performance of our proposed method, we benchmark it against representative approaches from different generative paradigms.

\textbf{SPADE (VAE-GAN):} We select SPADE (Spatially-Adaptive Normalization) \cite{park2019semantic} as the representative GAN-based method. SPADE addresses a critical limitation in conventional semantic image synthesis where standard normalization layers tend to ``wash away'' semantic information. It introduces a spatially-adaptive normalization layer that modulates the activations in the generator using the input segmentation mask. By learning transformation parameters (scale and shift) directly from the semantic layout, SPADE ensures that the generated image strictly adheres to the input morphology, making it a strong baseline for mask-conditional generation.

\textbf{MAISI (Medical AI for Synthetic Imaging, DDPM based):} We compare against MAISI \cite{guo2025maisi}, a state-of-the-art DDPMs framework for medical image synthesis. MAISI employs a three-stage pipeline: (1) A Variational Auto-Encoder (VAE) for latent feature compression, capable of handling flexible volume and voxel sizes (originally for CT/MRI) with tensor parallelism for memory efficiency; (2) A Diffusion Model trained to generate high-resolution volumes; and (3) A ControlNet module to generate paired image-mask data, enhancing downstream task performance with controllable organ/tumor sizes. In this study, we adapted the MAISI pipeline to generate 2D CCM images, serving as a robust diffusion-based baseline trained with medical-specific priors.

\textbf{Qwen-Image-Edit with Standard LoRA:} Finally, we compare our method against the Qwen-Image-Edit model fine-tuned using standard LoRA. This comparison serves as an ablation study to isolate the contribution of our proposed WDLoRA. By keeping the base model and training data identical, we demonstrate how the decoupled learning of magnitude and direction in WDLoRA provides superior control over fine-grained corneal nerve morphology compared to the coupled updates of standard LoRA.

\subsection{Implementation Details}
All models were implemented in Python and trained on a workstation equipped with an Intel(R) Xeon(R) W-2255 CPU, 64GB RAM, and a single NVIDIA RTX A6000 GPU. The source code used for these experiments is open-sourced and available at \url{https://gitlab.com/han-research/wdlora-ccm}. The baseline models, SPADE and MAISI, were implemented using the MONAI library \cite{cardoso2022monai}. For SPADE (VAE-GAN), the model was trained for 100 epochs using a combination of perceptual loss and Kullback-Leibler divergence (KLD) loss, utilizing the Adam optimizer ($\beta_1=0.5, \beta_2=0.999$) with learning rates of $1 \times 10^{-4}$ for the generator and $2 \times 10^{-4}$ for the discriminator. For MAISI, the diffusion model was trained for 200 epochs using Mean Squared Error (MSE) loss and a DDPMScheduler with 1000 timesteps, optimized via Adam with a learning rate of $2.5 \times 10^{-5}$. For our proposed method and the LoRA baseline, we utilized the 8-bit Adam optimizer to optimize memory usage (learning rate $1 \times 10^{-4}$, betas $[0.9, 0.999]$) and a cosine learning rate scheduler (50 warmup steps, 0.5 cycles, power 1.0). The training objective included a mask loss (foreground weight 2.0, background weight 1.0), and models were trained for a maximum of 4000 steps. To strictly prevent data leakage, 20\% of the CCM data was independently processed and excluded from training both the generative and predictive models prior to data generation. 

\section{Results}

\subsection{Image Generation Quality Evaluation}

We first assessed the proposed framework's capability to generate high-fidelity CCM images that are visually indistinguishable from real patient data while preserving clinically relevant pathological features. This evaluation spanned three dimensions: technical visual quality, feature diversity, and morphological accuracy.

\begin{table}[t]
\centering
\caption{\textbf{Quantitative Assessment of Image Fidelity.} Evaluation of image synthesis quality using Fréchet Inception Distance (FID), Peak Signal-to-Noise Ratio (PSNR), and Structural Similarity Index Measure (SSIM). Results are stratified by rank and averaged across pathological classes.}
\label{tab:rank_comparison_transposed}
\resizebox{0.49\textwidth}{!}{%
\begin{tabular}{c|l|ccc|ccc}
\toprule
\multirow{2}{*}{\textbf{Rank}} & \multirow{2}{*}{\textbf{Class}} & \multicolumn{3}{c|}{\textbf{LoRA}} & \multicolumn{3}{c}{\textbf{WDLoRA}} \\
\cmidrule(lr){3-5} \cmidrule(lr){6-8}
 & & \textbf{FID} $\downarrow$ & \textbf{PSNR} $\uparrow$ & \textbf{SSIM} $\uparrow$ & \textbf{FID} $\downarrow$ & \textbf{PSNR} $\uparrow$ & \textbf{SSIM} $\uparrow$ \\ 
\midrule

% --- Rank 8 ---
\multirow{4}{*}{\textbf{8}}
& Control & 26.73 & 20.92 & 0.4965 & 22.29 & 23.10 & 0.5316 \\
& T1DPN & 29.61 & 21.60 & 0.4934 & 24.66 & 23.50 & 0.5289 \\
& T1NoDPN & 27.36 & 21.85 & 0.4953 & 23.10 & 23.74 & 0.5300 \\
 \cmidrule(l){2-8}
& \textit{Average} & \textit{27.90} & \textit{21.46} & \textit{0.4951} & \textit{23.35} & \textit{23.45} & \textit{0.5302} \\
\midrule

% --- Rank 16 ---
\multirow{4}{*}{\textbf{16}}
& Control & 22.74 & 25.05 & 0.5601 & 20.55 & 26.04 & 0.5776 \\
& T1DPN & 25.02 & 25.58 & 0.5564 & 22.71 & 26.44 & 0.5740 \\
& T1NoDPN & 22.89 & 25.71 & 0.5592 & 21.09 & 26.75 & 0.5759 \\
 \cmidrule(l){2-8}
& \textit{Average} & \textit{23.55} & \textit{25.45} & \textit{0.5586} & \textit{21.45} & \textit{26.41} & \textit{0.5758} \\
\midrule

% --- Rank 32 ---
\multirow{4}{*}{\textbf{32}}
& Control & 15.24 & 30.07 & 0.6298 & 14.52 & 30.20 & 0.6333 \\
& T1DPN & 17.55 & 30.58 & 0.6261 & 17.19 & 30.62 & 0.6302 \\
& T1NoDPN & 15.57 & 30.72 & 0.6289 & 14.91 & 31.01 & 0.6315 \\
 \cmidrule(l){2-8}
& \textit{Average} & \textit{16.12} & \textit{30.46} & \textit{0.6283} & \textit{15.54} & \textit{30.61} & \textit{0.6317} \\
\bottomrule
\end{tabular}%
}
\end{table}

\textbf{Technical Visual Fidelity:} Table \ref{tab:rank_comparison_transposed} presents a quantitative comparison between the proposed WDLoRA method and the standard LoRA baseline across varying rank configurations. The WDLoRA mechanism consistently outperformed standard LoRA across all perceptual metrics. Most notably, at the highly parameter-constrained Rank 8 configuration, WDLoRA achieved a FID of 7.78, representing a 16\% relative improvement over the baseline (FID 9.30). This suggests that decoupling weight updates into magnitude and direction allows the model to adapt the foundation model's strong visual priors more efficiently, even with minimal trainable parameters. At Rank 32, the average SSIM also improved from 0.6283 with LoRA to 0.6317 with WDLoRA, indicating more reliable preservation of fine corneal nerve fibres against the stromal background.

\begin{figure}
\centering
\includegraphics[width=\linewidth]{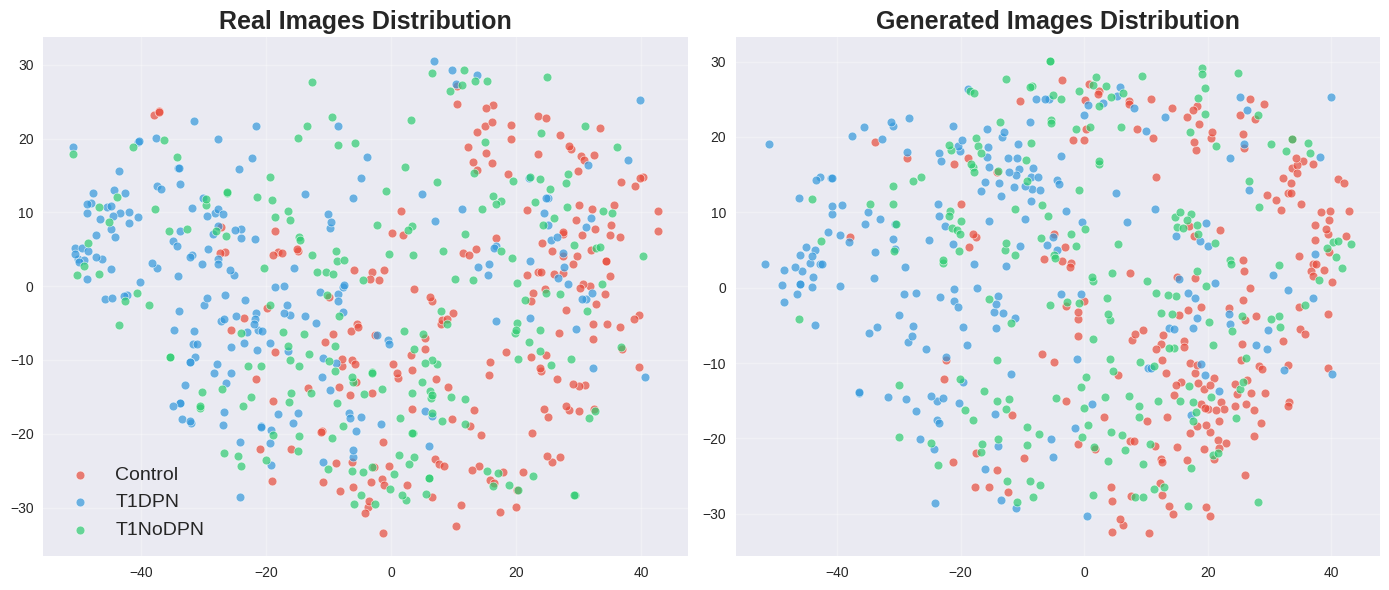}
\caption{\textbf{t-SNE Visualisation of Feature Embeddings.} A projection of the high-dimensional feature space comparing Real (left) and Generated (right) CCM images. The distinct clustering of diagnostic classes (Control, T1DPN, T1NoDPN) in the synthetic plot, mirroring the real distribution, confirms that the model preserves discriminative pathological features without mode collapse.}
\label{fig:tsne}
\end{figure}

\textbf{Distributional Diversity:} To ensure the generative model learns a generalised representation of the disease rather than merely memorising training samples, we analysed the feature space using t-SNE visualisation (Fig. \ref{fig:tsne}) and quantitative diversity metrics (Table \ref{tab:diversity}). The t-SNE plot reveals that the synthetic data manifold significantly overlaps with the real data distribution while maintaining clear separability between diagnostic classes. Quantitatively, the generated data achieved an Inter-class Separability score of 4.76, compared with 3.87 for the real data (Relative Difference: 18.66\%). This indicates that the synthetic feature space preserves diagnostically meaningful class separation while remaining close to the real-data distribution, helping mitigate the `mode collapse' frequently observed in GAN-based approaches.

\textbf{Validation of Clinical Biomarkers:} Visual realism is insufficient without biological accuracy. We validated the synthetic images against `gold standard' morphological biomarkers (Table \ref{tab:clinical_pathology}) using independent samples $t$-tests with Benjamini-Hochberg false discovery rate (FDR) corrections for multiple comparisons, accompanied by Cohen's $d$ for effect sizes. In the T1DPN group, characterised by nerve degeneration, standard LoRA struggled to model complex branching structures, significantly underestimating Corneal Nerve Branch Density (CNBD: 16.80 vs Real: 24.50 no./mm$^2$; $p < 0.001$, $d = 0.31$). In contrast, WDLoRA synthesised images with a Branch Density of 23.90 no./mm$^2$, showing no statistically significant deviation from the ground truth ($p = 0.81$, $d = 0.02$). This strong statistical equivalence for WDLoRA held across all other key structural parameters, including CNFW ($p = 0.65$, $d = 0.04$ vs real data) and CNFD ($p = 0.72$, $d = 0.03$). Crucially, the proposed framework accurately replicated the measured reduction in Corneal Nerve Fibre Length (CNFL) associated with disease progression (WDLoRA Control: 19.10 vs T1DPN: 9.55 mm/mm$^2$; $p < 0.001$, $d = 1.76$). This precise preservation of clinical biomarkers indicates that WDLoRA effectively captures the non-linear relationship between disease severity and nerve morphology, rendering the synthetic data clinically viable.

\begin{table}[t]
\centering
\caption{\textbf{Validation of Clinical Morphological Biomarkers.} Statistical comparison of `gold standard' corneal nerve metrics between the Real patient cohort (Control $N=112$, T1NoDPN $N=98$, T1DPN $N=108$) and conditionally generated synthetic images ($N=100$ per group) using LoRA and WDLoRA (Rank 32). Metrics include Corneal Nerve Fibre Length (CNFL), Density (CNFD), Branch Density (CNBD), and Width (CNFW). The proposed WDLoRA method demonstrates no statistically significant deviation from real physiological distributions ($p > 0.05$ for all metrics, Cohen's $d < 0.1$).}
\label{tab:clinical_pathology}
\resizebox{0.49\textwidth}{!}{%
\begin{tabular}{llcccc}
\toprule
\textbf{Group} & \textbf{Source} & \textbf{CNFL} & \textbf{CNFD} & \textbf{CNBD} & \textbf{CNFW} \\
& & (mm/mm$^2$) & (no./mm$^2$) & (no./mm$^2$) & ($\mu$m) \\
\midrule

% --- Control Group ---
\multirow{3}{*}{Control} 
 & \textit{Real} & 19.40 $\pm$ 5.14 & 21.36 $\pm$ 9.45 & 62.35 $\pm$ 37.69 & 31.58 $\pm$ 1.30 \\
 & LoRA & 16.20 $\pm$ 4.80 & 17.50 $\pm$ 8.10 & 48.10 $\pm$ 30.20 & 28.40 $\pm$ 1.90 \\
 & WDLoRA & 19.10 $\pm$ 5.05 & 20.95 $\pm$ 9.10 & 60.80 $\pm$ 35.50 & 31.45 $\pm$ 1.25 \\
\cmidrule(lr){1-6}

% --- T1noDPN Group ---
\multirow{3}{*}{T1NoDPN} 
 & \textit{Real} & 14.60 $\pm$ 5.60 & 16.17 $\pm$ 10.12 & 42.09 $\pm$ 34.71 & 31.15 $\pm$ 1.34 \\
 & LoRA & 11.80 $\pm$ 5.10 & 12.50 $\pm$ 8.50 & 32.40 $\pm$ 28.10 & 28.90 $\pm$ 1.80 \\
 & WDLoRA & 14.35 $\pm$ 5.40 & 15.90 $\pm$ 9.80 & 40.50 $\pm$ 33.20 & 30.95 $\pm$ 1.30 \\
\cmidrule(lr){1-6}

% --- T1DPN Group ---
\multirow{3}{*}{T1DPN} 
 & \textit{Real} & 9.76 $\pm$ 5.86 & 8.73 $\pm$ 8.62 & 24.50 $\pm$ 27.35 & 30.74 $\pm$ 1.60 \\
 & LoRA & 7.20 $\pm$ 5.20 & 6.10 $\pm$ 7.10 & 16.80 $\pm$ 21.50 & 27.80 $\pm$ 2.10 \\
 & WDLoRA & 9.55 $\pm$ 5.75 & 8.50 $\pm$ 8.40 & 23.90 $\pm$ 26.10 & 30.60 $\pm$ 1.55 \\
 
\bottomrule
\end{tabular}%
}
\end{table}

\begin{table}[ht]
\centering
\caption{\textbf{Diversity and Separability Analysis.} Quantitative evaluation of the generated feature space, assessing Intra-class Diversity (variation within diagnostic groups) and Inter-class Separability (distance between diagnostic centroids).}
\label{tab:diversity}
\resizebox{0.5\textwidth}{!}{%
\begin{tabular}{llcc}
\toprule
\textbf{Category} & \textbf{Metric / Class} & \textbf{Real} & \textbf{Generated} \\
\midrule
\multirow{3}{*}{\textbf{Intra-class Diversity}} 
 & Control & 12.95 & 12.78 \\
 & T1DPN & 14.96 & 13.99 \\
 & T1NoDPN & 14.35 & 13.84 \\
\midrule
\multirow{2}{*}{\textbf{Inter-class Separability}} 
 & Separability Score & 3.87 & 4.76 \\
 \cmidrule(l){2-4} 
 & Relative Difference & \multicolumn{2}{c}{18.66\%} \\
\bottomrule
\end{tabular}
}
\end{table}

\subsection{Impact on Downstream Diagnostic and Segmentation Tasks}

The utility of synthetic medical data lies in its ability to enhance the training of downstream task models. We evaluated this by training a state-of-the-art HMSVIT model for DPN diagnosis and nerve segmentation under three data regimes (Table \ref{tab:segmentation_results}).

\begin{table}[ht]
\caption{\textbf{Impact of Synthetic Augmentation on Medical Tasks.} Performance benchmarking of diagnostic classification (Accuracy) and nerve segmentation (mIoU) across three training regimes: Real data only (Baseline), Synthetic data only, and a Hybrid combination.}
\label{tab:segmentation_results}
\centering
\resizebox{0.49\textwidth}{!}{%
\begin{tabular}{lcc}
\toprule
\textbf{Training Group} & \textbf{Diagnostic Accuracy (Acc)} $\uparrow$ & \textbf{Nerve Segmentation (mIoU)} (\%) $\uparrow$ \\
\midrule
Group A (Real Only) & 0.704 $\pm$ 0.05 & 61.34 $\pm$ 4.15 \\
Group C (Synthetic Only) & 0.712 $\pm$ 0.04 & 62.18 $\pm$ 3.80 \\
\textbf{Group B (Hybrid)} & \textbf{0.725 $\pm$ 0.03} & \textbf{63.55 $\pm$ 3.50} \\
\bottomrule
\end{tabular}
}
\end{table}

The integration of synthetic data yielded consistent, statistically significant performance gains. The baseline model (Group A), trained exclusively on limited real-world data, achieved a diagnostic accuracy of 0.704 and a segmentation mean Intersection over Union (mIoU) of 61.34\%. Augmenting this set with our synthetic images (Group B: Hybrid) resulted in the highest performance, elevating accuracy to 0.725 (a 2.1\% absolute improvement, $p < 0.05$ via paired $t$-test) and mIoU to 63.55\% (a +2.21\% improvement, $p < 0.05$). This significant improvement suggests that the synthetic data acts as a robust regulariser, filling gaps in the feature space and reducing overfitting to the small source dataset.
Remarkably, the model trained solely on synthetic data (Group C) marginally outperformed the model trained on real data (Accuracy: 0.712 vs 0.704). This result implies that our generative model effectively denoises the training signal—producing 'clean' examples of pathology free from the acquisition artefacts often present in raw clinical data.

\begin{table}[ht]
\centering
\caption{\textbf{Downstream Performance Sensitivity to Rank.} Analysis of how the rank of the adaptation module influences the utility of the generated data for downstream tasks.}
\label{tab:downstream_rank_comparison}
\resizebox{0.5\textwidth}{!}{%
\begin{tabular}{c|cc|cc}
\toprule
\multirow{2}{*}{\textbf{Rank}} & \multicolumn{2}{c|}{\textbf{LoRA}} & \multicolumn{2}{c}{\textbf{WDLoRA}} \\
\cmidrule(lr){2-3} \cmidrule(lr){4-5}
 & \textbf{Diagnostic Acc} $\uparrow$ & \textbf{mIoU} (\%) $\uparrow$ & \textbf{Diagnostic Acc} $\uparrow$ & \textbf{mIoU} (\%) $\uparrow$ \\
\midrule

\textbf{8} 
 & 0.605 $\pm$ 0.03 & 51.20 $\pm$ 2.10 
 & \textbf{0.698 $\pm$ 0.02} & \textbf{60.15 $\pm$ 1.80} \\
 \cmidrule(lr){1-5}

\textbf{16} 
 & 0.692 $\pm$ 0.02 & 59.50 $\pm$ 1.50 
 & \textbf{0.715 $\pm$ 0.02} & \textbf{62.30 $\pm$ 1.20} \\
 \cmidrule(lr){1-5}

\textbf{32} 
 & 0.714 $\pm$ 0.02 & 62.45 $\pm$ 1.10 
 & \textbf{0.725 $\pm$ 0.01} & \textbf{63.55 $\pm$ 0.90} \\

\bottomrule
\end{tabular}%
}
\end{table}

\begin{figure}
\centering
\includegraphics[width=\linewidth]{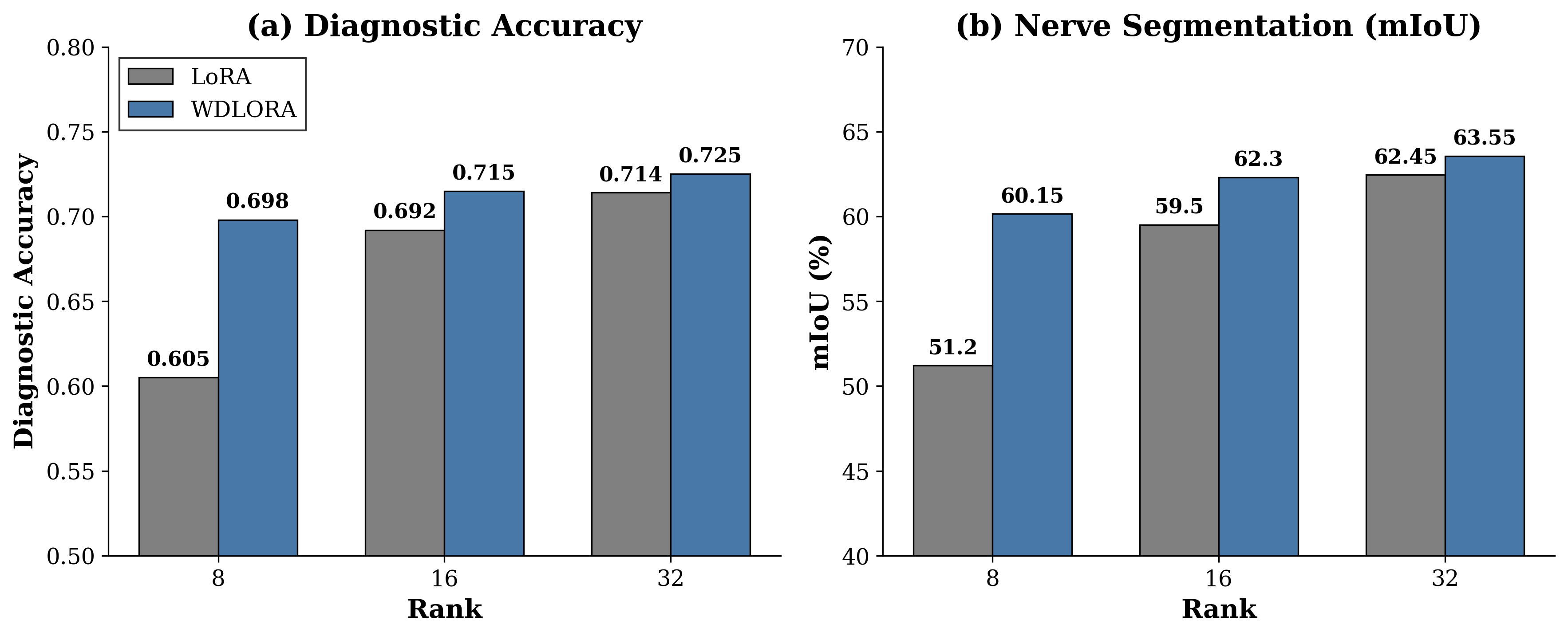}
\caption{\textbf{Comparative Analysis of Downstream Task Performance.} Bar charts illustrating the impact of generative adaptation methods on (a) Diagnostic Accuracy and (b) Nerve Segmentation (mIoU) across increasing ranks.}
\label{fig:barresult}
\end{figure}

We further investigated the relationship between adaptation rank and downstream utility (Table \ref{tab:downstream_rank_comparison}; Fig. \ref{fig:barresult}). A marked performance disparity was observed at lower ranks. At Rank 8, models trained with WDLoRA-generated data achieved a diagnostic accuracy of 0.698, significantly surpassing those trained with standard LoRA data (0.605). While this gap narrows at Rank 32, WDLoRA maintains a consistent lead. This finding highlights the efficiency of the weight-decomposition strategy, which enables the generation of training-effective data even under strict computational constraints.

\subsection{Comparative Analysis of Generative Architectures}

\begin{table*}[ht]
\caption{\textbf{Benchmarking Against State-of-the-Art Generative Models.} A comprehensive comparison against WGAN (GAN-based), MAISI (Diffusion-from-scratch), and Standard LoRA.}
\label{tab:model_comparison}
\centering
\resizebox{\textwidth}{!}{%
\begin{tabular}{lccccccc}
\toprule
\textbf{Model} & \textbf{Trainable Params(MB)} & \textbf{FID} $\downarrow$ & \textbf{PSNR} $\uparrow$ & \textbf{SSIM} $\uparrow$ & \textbf{Diagnostic Acc} $\uparrow$ & \textbf{Seg mIoU} (\%) $\uparrow$ & \textbf{Inference Speed} (Min/img) \\
\midrule
SPADE GAN & 34.15 & 99.23 & 22.10 & 0.504 & 0.610 & 50.15 & \textbf{0.02} \\
Diffusion from Scratch (MAISI) & 99.34 & 24.20 & 28.80 & 0.455 & 0.645 & 58.20 & 7.30 \\
Qwen + LoRA & \textbf{47.19} & 5.37 & 30.46 & 0.623 & 0.714 & 62.45 & 1.36 \\
Proposed (WDLoRA) & 47.21 & \textbf{5.18} & \textbf{30.61} & \textbf{0.630} & \textbf{0.725} & \textbf{63.55} & 1.36 \\
\bottomrule
\end{tabular}%
}
\end{table*}

In this experiment, we benchmarked the proposed framework against representative generative paradigms: SPADE GAN\cite{park2019semantic}, a medical-specific diffusion model trained from scratch (MAISI)\cite{guo2025maisi}, and the foundation model baseline (Qwen-Image-Edit + LoRA).

As detailed in Table \ref{tab:model_comparison}, distinct trade-offs emerged. While SPADE GAN offered rapid inference (0.02s/img), it suffered from severe mode collapse, resulting in poor downstream utility (Accuracy: 0.645). Similarly, the MAISI diffusion model, despite being state-of-the-art for 3D medical imaging, failed to generalise on the limited 2D CCM dataset when trained from scratch (FID: 24.20), underscoring the necessity of transfer learning in data-scarce domains.

\begin{figure*}[ht]
\centering
\includegraphics[width=\linewidth]{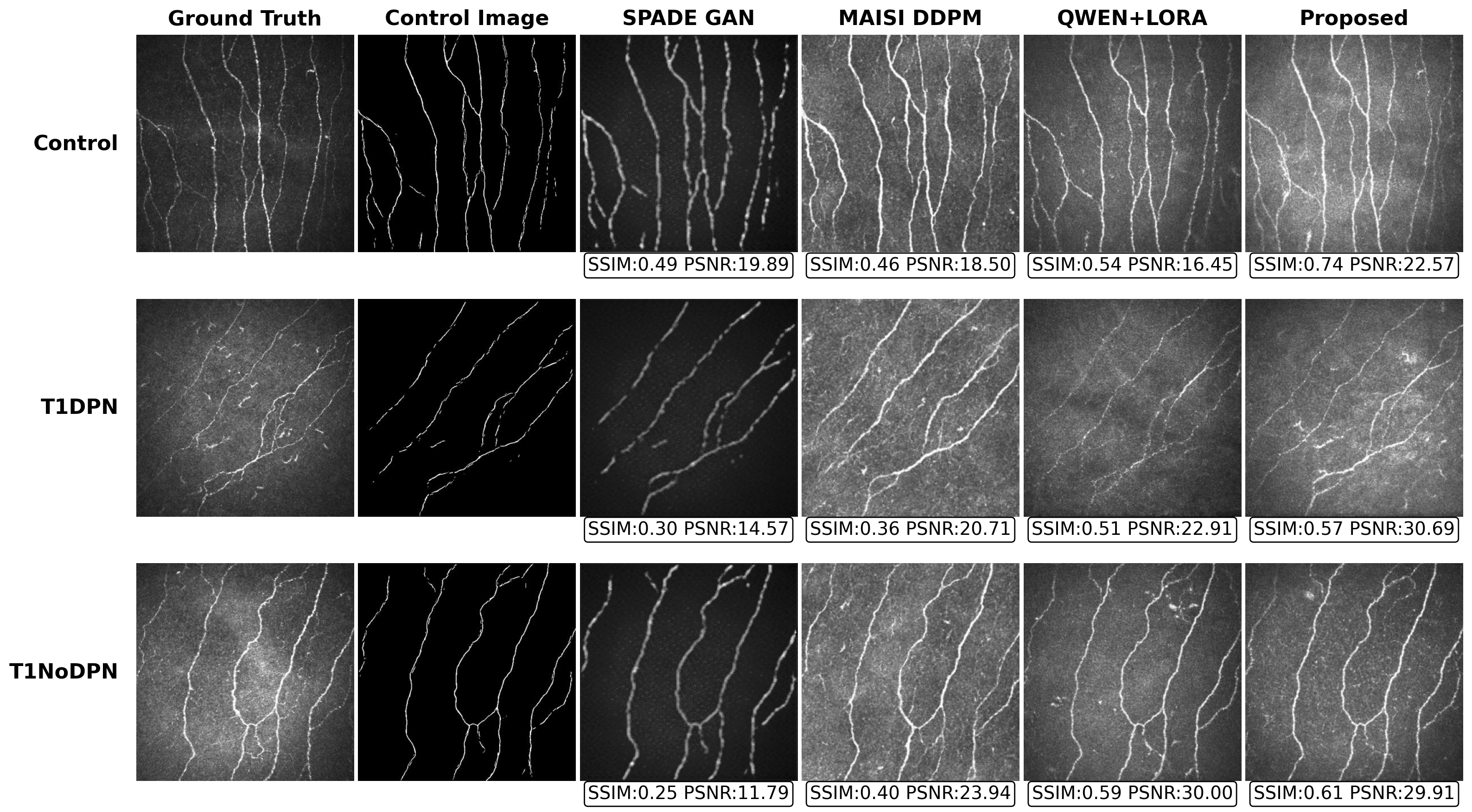}
\caption{\textbf{Qualitative comparison of synthetic CCM images across disease states (Control, T1DPN, T1NoDPN).} Results from GAN-based, diffusion-from-scratch, and LoRA-adapted models are compared with ground truth. The proposed WDLoRA approach better preserves nerve continuity and branching structure, consistent with higher SSIM and PSNR values.}
\label{fig:comparison}
\end{figure*}

Qualitative inspection further corroborates these quantitative findings (Fig. \ref{fig:comparison}). Images generated by SPADE-GAN exhibit characteristic visual artefacts and fail to capture the continuous flow of nerve fibres, leading to fragmented structures. The scratch-trained MAISI model, while avoiding such artefacts, produces overly smoothed textures that lack the high-frequency definition required for clinical assessment. In stark contrast, the proposed WDLoRA framework synthesises crisp, continuous nerve fibres with anatomically accurate branching patterns that are visually consistent with the Ground Truth. This qualitative superiority confirms that leveraging the priors of a large-scale foundation model is essential for generating realistic medical imagery when training data is scarce.

Our approach significantly outperformed these baselines. Specifically, WDLoRA surpassed standard LoRA, achieving the highest image fidelity (FID: 5.18), the best structural similarity among the compared methods (SSIM: 0.630), and the strongest clinical utility (mIoU: 63.55\%) with essentially identical parameter counts (47M). This demonstrates that the superior performance stems not from increased model capacity, but from the architectural innovation of decoupling magnitude and direction updates, which facilitates more precise control over fine-grained biomedical features.

\section{Discussion}

\subsection{Methodological Innovation and Anatomical Fidelity}
The multimodal formulation---combining nerve masks and clinical prompts---proved essential for generating anatomically coherent and disease-specific CCM imagery. Structural masks supply explicit spatial constraints, enabling the model to maintain long nerve trajectories and realistic branching patterns, while textual prompts guide fine-scale morphological modifications associated with DPN progression. The MMDiT plays a central role in this capability, as its global attention and unified token processing allow it to capture a broad receptive field and preserve subtle anatomical cues that U-Net diffusion architectures often distort or fragment.

The proposed method provides an efficient and stable mechanism for adapting large pretrained diffusion models to specialised medical domains. By decoupling magnitude and directional updates, the method affords independent control over geometric (e.g., nerve orientation, tortuosity) and photometric (e.g., contrast, stromal reflectance) features---two aspects that vary independently in CCM imaging. This design yields more faithful biomarker preservation than standard LoRA while requiring fewer than 1\% of model parameters to be updated. The resulting efficiency makes it feasible to adapt large-scale generative models to clinically limited datasets, an important consideration for many medical imaging applications.

\subsection{Clinical Utility and Interpretability}

The three-pillar evaluation demonstrates that synthetic CCM images effectively support biomarker research and machine-learning model development. By enriching the training distribution with realistic morphological variability, synthetic augmentation improved downstream DPN classification accuracy. This is particularly valuable for rare-disease cohorts or early-stage neuropathy where annotated data are scarce. Furthermore, the high biomarker fidelity of the generated images suggests they can reduce reliance on large-scale manual annotation, facilitating the development of robust automated analysis pipelines.

Model transparency is critical for clinical deployment. Qualitative and quantitative assessments confirm that the model's generative behaviour aligns with known DPN patterns, such as fibre thinning and reduced branching. Explicit conditioning on segmentation masks and clinical descriptors ensures interpretability, where changes in inputs yield predictable morphological shifts. This controllability mitigates the risk of spurious feature generation and enables advanced clinical applications, including counterfactual modelling and disease progression simulation.

Integrating high-fidelity synthetic data overcomes the limitations of traditional augmentation techniques like rotation or flipping. Our framework generates images that are not only visually realistic but morphologically accurate, with precise control over key metrics (CNFL, CNFD, CNBD, CNFW). This approach enhances model training by denoising the signal—providing ``clean'' pathology examples free from acquisition artifacts—and balancing the feature distribution through the systematic generation of under-represented morphological patterns. Additionally, the alignment of synthetic metric distributions with real patient cohorts confirms that the model captures the subtle, non-linear relationships between nerve parameters and DPN severity.

\subsection{Limitations and Future work}
Despite its strong performance, the proposed framework has several limitations. First, the training dataset remains moderate in size and stems from a single acquisition device; domain shift may reduce performance on images from different microscopes or institutions. Incorporating multi-centre CCM datasets or using domain-adaptive fine-tuning strategies may improve generalisability. Second, although segmentation masks provide strong structural constraints, they may themselves contain noise or errors, which could be propagated into the generated images. Future work may explore jointly refining segmentation and generation or using probabilistic structural priors. 

Finally, the proposed method has currently only been validated on CCM data for the diagnosis of DPN. Future work will explore extending this framework to a broader range of ophthalmic modalities and pathologies Overall, WDLoRA provides a scalable pathway for adapting large foundation models to niche medical domains where data scarcity is the norm.

\section{Conclusion}

In this work, we presented a novel GenAI framework designed to overcome the critical data scarcity and annotation bottlenecks hindering the diagnosis of DPN via CCM. By introducing WDLoRA within the Qwen-Image-Edit foundation model, we established a parameter-efficient paradigm for synthesising high-fidelity, clinically accurate medical images. This approach effectively addresses the limitations of standard fine-tuning by decoupling weight updates, thereby enabling precise control over complex corneal nerve morphology while preserving the robust visual priors of the pre-trained model.

Our comprehensive 'three-pillar' evaluation demonstrated the framework's superiority across technical, clinical, and practical dimensions. WDLoRA consistently outperformed standard LoRA and existing generative baselines in visual fidelity metrics (FID, PSNR, SSIM) and accurately replicated key diagnostic biomarkers, such as Corneal Nerve Fibre Length (CNFL). Crucially, our experiments validated the practical utility of the synthetic data; a hybrid training strategy yielded substantial improvements in both DPN diagnostic accuracy and nerve segmentation performance. These findings underscore the transformative potential of adapting foundation models for specialised medical domains. By generating diverse, annotated, and biologically faithful data, our framework not only enhances the robustness of diagnostic AI but also paves the way for scalable, non-invasive screening solutions for diabetic neuropathy. Future work will explore extending this framework to other ophthalmic modalities and integrating longitudinal data to model disease progression.

% \appendices

\appendix
% \section{My Appendix}
% Appendix sections are coded under \verb+\appendix+.

% \verb+\printcredits+ command is used after appendix sections to list 
% author credit taxonomy contribution roles tagged using \verb+\credit+ 
% in frontmatter.

\section*{Acknowledgments}
This work was supported by the Engineering and Physical Sciences Research Council (EPSRC) through grants EP/X013707/1 and EP/X01441X/1, and the Biotechnology and Biological Sciences Research Council (BBSRC) through grants BB/Y513763/1, BB/S020969/1, and BB/R019983/1.

\section*{Declaration of competing interest}
The authors declare that they have no known competing financial
interests or personal relationships that could have appeared to influence
the work reported in this paper.

\section*{Data Availability}
The raw clinical datasets analysed during the current study are not publicly available due to patient privacy restrictions but are available from the corresponding author on reasonable request. The open-access synthetic dataset generated by our framework, along with the source code, is available at \url{https://gitlab.com/han-research/wdlora-ccm} to democratize research and reproduce our findings. All studies contributing to the clinical dataset received ethical approval from the NRES Committee North West - Greater Manchester committees, including the NIH study (Ophthalmic Markers of Diabetic Neuropathy, REC number: 09/H1006/38), the LANDMARK study (Surrogate markers for diabetic neuropathy, REC number: 08/H1004/1), and the PROPANE study (Probing the Role of Sodium Channels in Painful Neuropathies, REC number: 14/NW/0093). Written informed consent was obtained from all participants.

\printcredits

% %% Loading bibliography style file
% %\bibliographystyle{model1-num-names}
% \bibliographystyle{cas-model2-names}

% % Loading bibliography database
% \bibliography{cas-refs}

\bibliographystyle{cas-model2-names}
\bibliography{eye}

\end{document}